\documentclass[11pt]{article}

\usepackage{palatino}
\usepackage{amsmath}
\usepackage{amssymb}
\usepackage{natbib}
\usepackage{graphicx}

\usepackage{geometry}
\newif\ifarxiv
\newif\ifnotarxiv

\notarxivfalse
\arxivtrue

\usepackage{hyperref}

\usepackage{longtable}%

\usepackage{booktabs}
\usepackage[load-configurations=version-1]{siunitx} %
 \usepackage{floatrow}
 \usepackage{xspace}
\usepackage[table]{xcolor}
\date{}
\usepackage{url}

\newcommand{\matthijs}[1]{{\color{blue}[\textbf{Matthijs}:#1]}}

\newcommand{\teamname}[1]{\textsc{#1}}

\newcommand{\eg}{\emph{e.g.}\@\xspace}
\newcommand{\etc}{\emph{etc.}\@\xspace}
\newcommand{\ie}{\emph{i.e.}\@\xspace}

\newcommand{\wrt}{\emph{w.r.t.}\@\xspace}

\newcommand{\mta}{{\textsc{VisionForce}{\footnotesize-mt1}}\@\xspace}
\newcommand{\mtb}{{\textsc{Separate}{\footnotesize-mt2}}\@\xspace}
\newcommand{\mtc}{{\textsc{ImgFp}{\footnotesize-mt3}}\@\xspace}
\newcommand{\dta}{{\textsc{Lyakaap}{\footnotesize-dt1}}\@\xspace}
\newcommand{\dtb}{{\textsc{S-squared}{\footnotesize-dt2}}\@\xspace}
\newcommand{\dtc}{{\textsc{VisionForce}{\footnotesize-dt3}}\@\xspace}

\title{Results and findings of the 2021 Image Similarity Challenge}

\newcommand{\aff}[1]{\textsuperscript{#1}}

\author{
\begin{minipage}{12cm} 
\bf
\centering
  Zo\"e Papakipos\aff{1},
  Giorgos Tolias\aff{2},
  Tomas Jenicek\aff{2},
  Ed Pizzi\aff{1},
  Shuhei Yokoo\aff{3},
  Wenhao Wang\aff{4}, Yifan Sun\aff{4}, Weipu Zhang\aff{4}, \\ Yi Yang\aff{4},
  Sanjay Addicam\aff{5}, Sergio Manuel Papadakis\aff{6},
  Cristian Canton Ferrer\aff{1},
  Ondřej Chum\aff{2},
  Matthijs Douze\aff{1} \\
  \rm \small 
  ~\\
  \aff{1} Meta AI, 
  \aff{2} Czech Technical University in Prague, 
  \aff{3} DeNA, Japan, 
  \aff{4} Baidu Research, 
  \aff{5} Chandler, Arizona USA, 
  \aff{6} San Carlos de Bariloche, Argentina
  \end{minipage} 
}

\begin{document}

\maketitle

\begin{abstract}
The 2021 Image Similarity Challenge introduced a dataset to serve as a new benchmark to evaluate recent image copy detection methods.
There were 200 participants to the competition.
This paper presents a quantitative and qualitative analysis of the top submissions.
It appears that the most difficult image transformations involve either severe image crops or hiding into unrelated images, combined with local pixel perturbations. 
The key algorithmic elements in the winning submissions are:
training on strong augmentations, self-supervised learning, score normalization, explicit overlay detection, and global descriptor matching followed by pairwise image comparison\footnote{This is the long version of a paper to appear in PMLR about the challenge.}
\end{abstract}

\section{Introduction}
\label{sec:intro} 

The Image Similarity Challenge organized in 2021  aimed to assess the efficacy of image copy detection algorithms using a large dataset with robust image edits. 
The challenge design aimed to reflect practical requirements for large-scale copy detection systems, where most queries do not match references in the dataset, and it is important to efficiently separate copies from non-copies.

The challenge consisted of two tracks: a descriptor track, and an unconstrained matching track. In the descriptor track, participants provide descriptor vectors in $\mathcal{R}^{256}$ for each image in the dataset, and matching is performed using L2 distances between the vectors. In the matching track, any matching techniques can be used, including  pairwise comparisons.
The challenge organizers created and released the DISC21 dataset for use in this challenge and beyond, as a benchmark for image copy detection. 
DISC21 includes a large reference set of images and a smaller set of query images, where the goal is to find the subset of matches between the two. 
See~\citep{douze20212021} for more details about the creation of the dataset.
The challenge drew over 200 participants in its final phase, including strong solutions. 
The main takeaways for image copy detection from this challenge include the following. 
(1) Strong and non-standard image augmentations that mimic typical cases of image copies are very beneficial in the training.
(2) Self-supervised learning by instance-discrimination is crucial, not only for pre-training, but also as the main training stage.
(3) Score normalization, either explicitly in the matching track, or implicitly by descriptor processing in the descriptor track, has a significant impact.
(4) Explicit overlay detection is a task-tailored approach that has proven useful.
(5) The use of regional representation and matching is able to significantly improve copy detection performance compared to global descriptor approaches.

This paper presents the results and methods from participants as well as some analysis of the solutions and dataset. 
\ifarxiv
The paper is organized in chronological order from the challenge organizer's point of view:
we first recap how the challenge was organized in Section~\ref{sec:thechallenge}. 
\fi
Section~\ref{sec:results} is an in-depth analysis of the results that were submitted. 
Section~\ref{sec:methods} describes what components the most successful participants used for their winning entries.
\if

\section{The ISC challenge}
\label{sec:thechallenge} 

The DISC21 dataset is described in detail in \citep{douze20212021}.
The dataset and baseline implementations are publicly available~\footnote{See~\url{https://sites.google.com/view/isc2021/dataset}, \url{https://github.com/facebookresearch/isc2021}}.
This is a summary of its structure. 
It includes a reference set of 1 million images, a development set of 50,000 augmented query images (a subset of which are transformed copies of a reference image), and a training set of 1 million images. 
About 60\% of the query images in DISC21 have been transformed using image augmentations from the AugLy library~\citep{augly}.
The remaining 40\% have been manually edited by humans.

During Phase 1 of the challenge, participants could transform and train their solution on the training set and evaluate and iterate their solution based on the matching accuracy of the development queries to the references. %
We released half of the ground truth for half of the queries, so participants could compute their accuracy on those 25,000 directly. To compute the accuracy on the full query set, participants could submit their solution to the challenge platform, DrivenData\footnote{\url{https://www.drivendata.org/competitions/84/}}.

Phase 2 of the challenge, the final evaluation to determine the winners, took place in October. We released an additional 50,000 test query images which included held-out transformations not seen in the development set, in order to test the robustness of participants' solutions. 
Being robust to data manipulations not seen at training time is crucial for an image copy detection system at scale, as users continually find new ways to manipulate data for both benign (\eg new meme formats or Instagram filters) and adversarial (\eg overlaying an image containing inappropriate content onto a benign background image to try to evade content moderation) reasons.

\section{Analysis of the results}
\label{sec:results} 

This section takes an outside view of the results without any insight into the methods used by the participants.
The following analysis is based on the raw submission files to the final track. 
It includes results from participants that were disqualified (\eg because they broke participation rules). 
For most results we included only the top submissions to improve the readability.

\subsection{High-level comparisons} 

\begin{figure}
\centering
\includegraphics[width=0.48\linewidth]{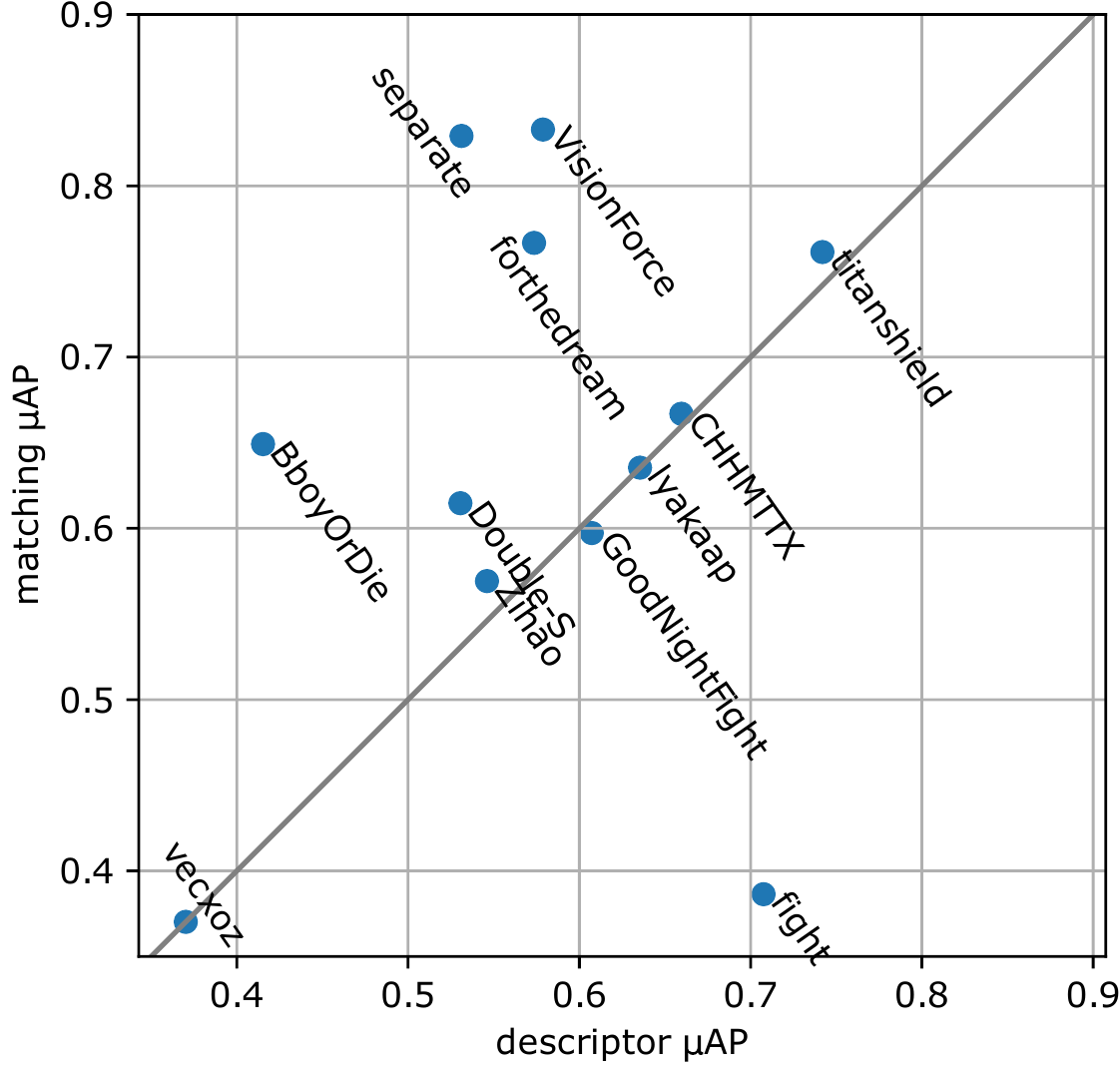}
\hfill
\includegraphics[width=0.48\linewidth]{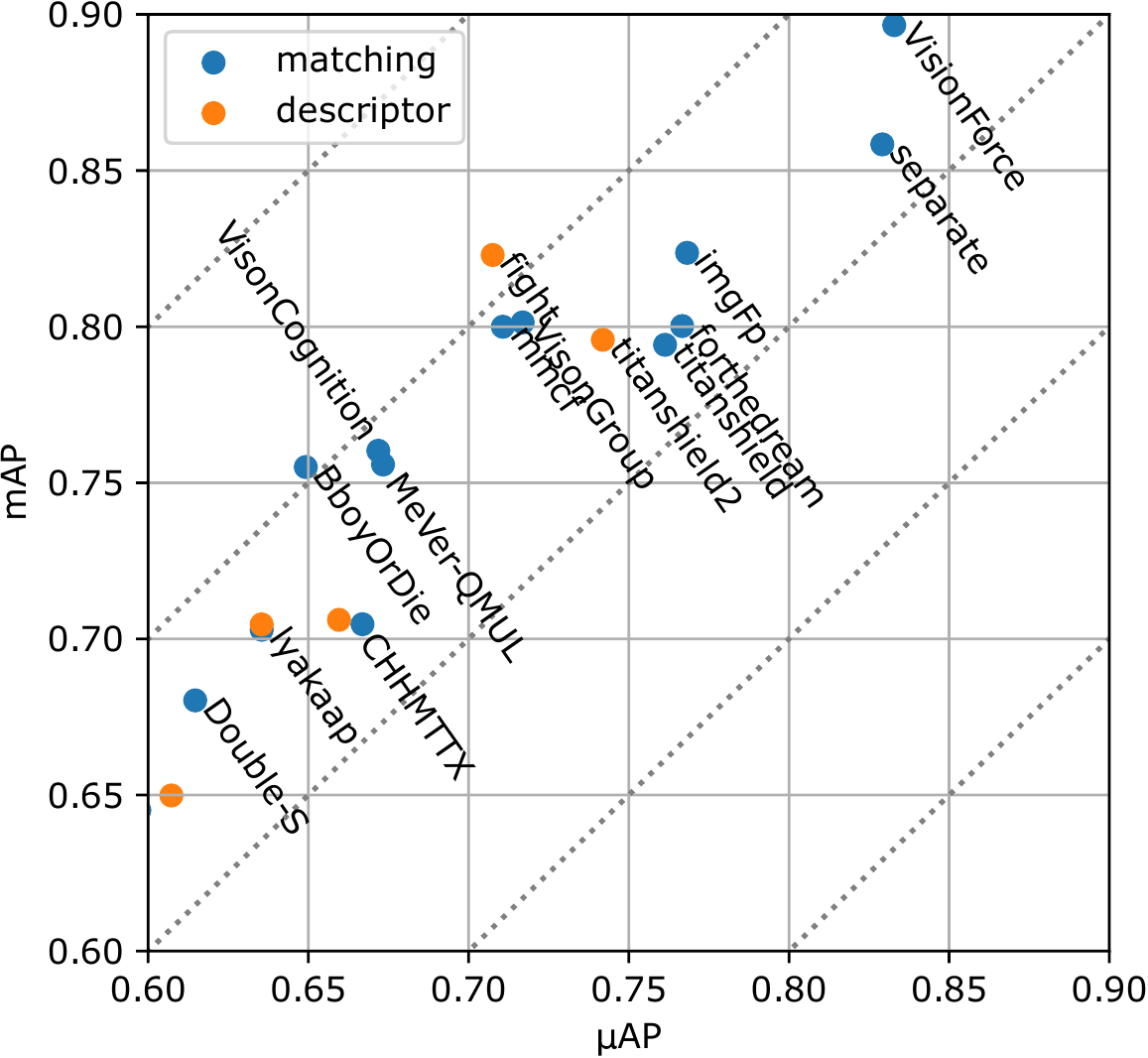}
\caption{\label{fig:highlevelcompare}
    Comparison of the ISC submissions on two axes: 
    Left: descriptor vs. matching track performance, 
    Right: $\mu$AP vs. mean AP (mAP $n=10,000$). 
}
\end{figure} 

The matching track is less constrained than the descriptor track; in fact a valid descriptor submission can be converted into a matching submission. Most participants to the descriptor track submitted to the matching track as well.
We compare the submissions to assess the performance gain by the matching. 
Figure~\ref{fig:highlevelcompare} (left) shows a maximum performance difference between the two track equal to 0.2 and almost equal performance for some submissions like teams~\teamname{titanshield}, \teamname{CHHMTX}, \teamname{lyakaap}. There is one outlier case where the descriptor submission is better than the matching one.                       

\paragraph{Per-query comparison.} 

The default evaluation metric, namely $\mu$AP, considers all queries jointly. We provide a finer analysis of the results, with the following per-query measure of performance. We consider only the 10k queries that actually match one of the reference images and discard the 40k distractors. 
We computre the \emph{average precision} for each query, which coincides with the inverse of the rank of the true positive result. 
Averaging this measure over the 10k queries results in the so-called mean average precision (mAP). 
The mAP can be computed on subsets of queries, in which case we indicate $n$, the number of query images of the subset.

Figure~\ref{fig:highlevelcompare} (right) shows how mAP compares to $\mu$AP per submission. 
The two measures are not directly comparable, but a large gap between mAP and $\mu$AP is a sign of an ineffective score normalization; $\mu$AP is designed to evaluate how well matching scores are normalized across queries, see~\cite[Section 4.4]{douze20212021}. 
For example, it is possible that for the matching track, the~\teamname{VisionForce} team's score normalization is worse than that of ~\teamname{separate}.

\subsection{Analysis per data source}

The DISC21 dataset is built from two different sources and uses two different ways to apply the transformations.

\begin{figure}
\centering
\includegraphics[width=0.48\linewidth]{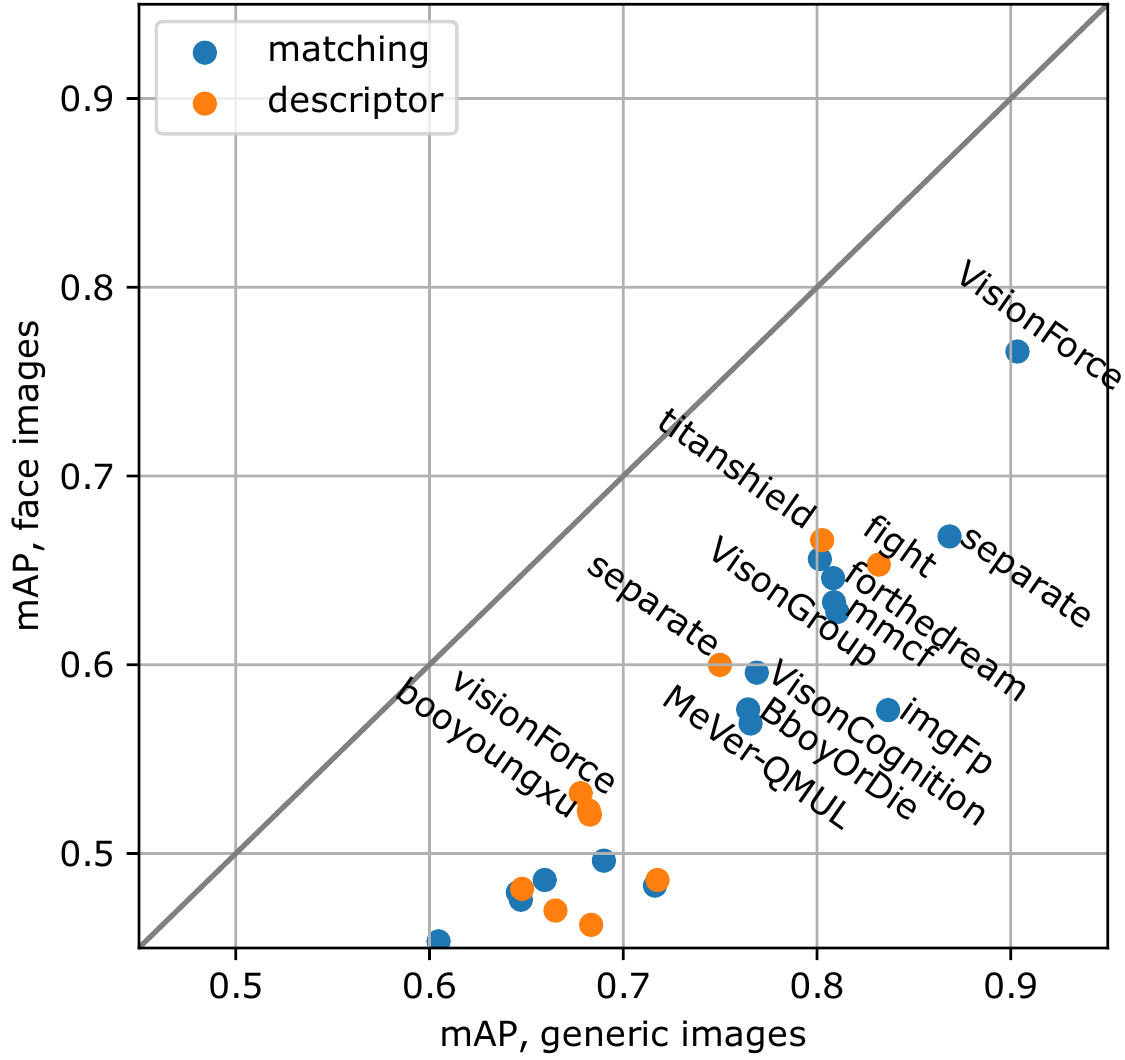}
\hfil
\includegraphics[width=0.48\linewidth]{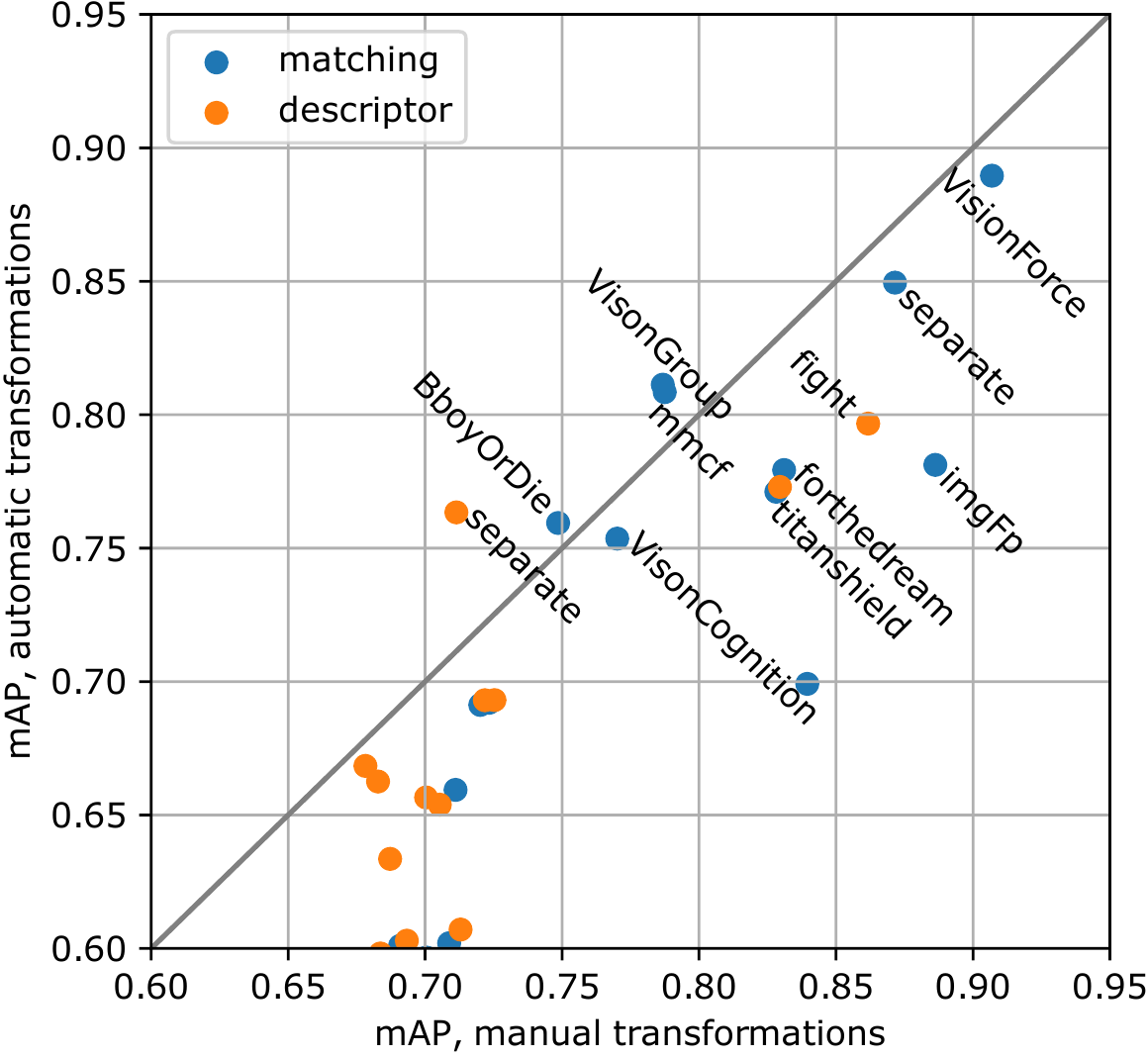}
\caption{\label{fig:breakdownpersource}
    Results per broad data source. 
    Left: reference images are face images (number of query images $n$=500) or generic images ($n$=9500), 
    Right: transformations performed manually ($n$=4040) or automatically ($n$=5960).
}
\end{figure}

\paragraph{Face images vs. generic images.}

DISC21 is built from two data sources, namely generic images from YFCC100M~\citep{Thomee2016YFCC100MTN} that contain no images of people and 5\% of face images from the DFDC challenge~\citep{DFDC2020}. 
The plot in Figure~\ref{fig:breakdownpersource} (left) shows that the scores for queries of face images is lower than that for generic images for all submissions. 
Figure~\ref{fig:exampleFPs} shows that false positive results often depict the same object/face from a sightly different viewpoint, which is a mismatch from a copy detection point of view and forms a very challenging case.  There are more pairs of images with such small variations in the images of faces, and they are more likely to be returned as results.

\newcommand{\discimage}[1]{\includegraphics[height=2cm]{figs/disc_images/#1}}
\newcommand{\imcredsF}[1]{\scalebox{0.4}{\rotatebox{90}{Flickr / #1}}}
\newcommand{\imcredsD}[1]{\scalebox{0.4}{\rotatebox{90}{DFDC image #1}}}

\begin{figure}
    \centering
    \begin{tabular}{cc@{}c|cc@{}c}
    \multicolumn{2}{c}{rank 7083} & 
    \multicolumn{2}{c}{rank 6980} \\
    \discimage{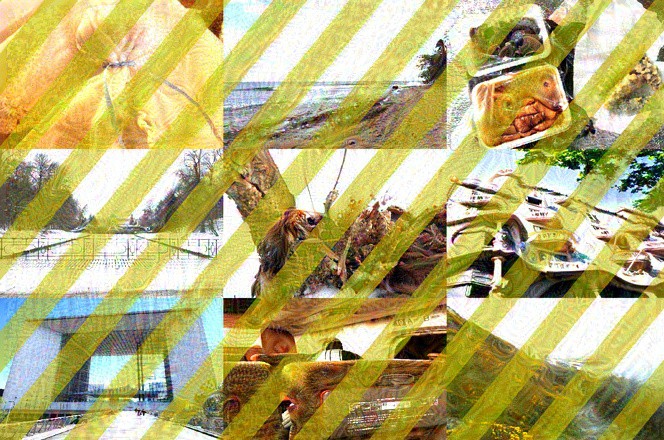} & \discimage{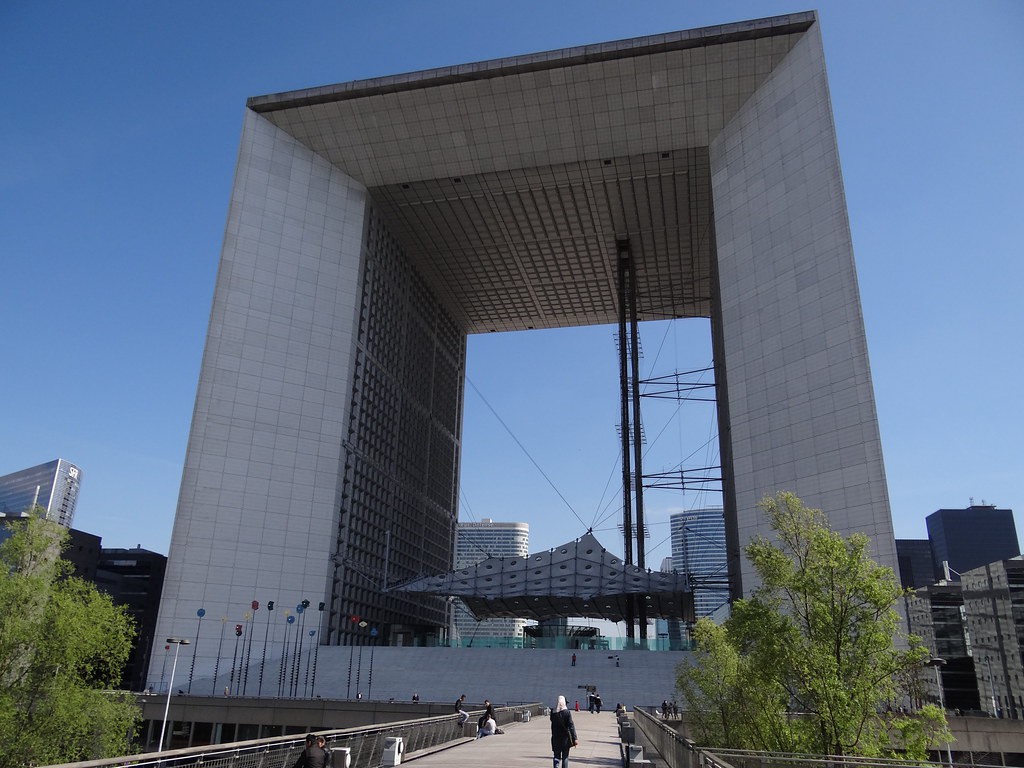} & \imcredsF{Flikkesteph} &
    \discimage{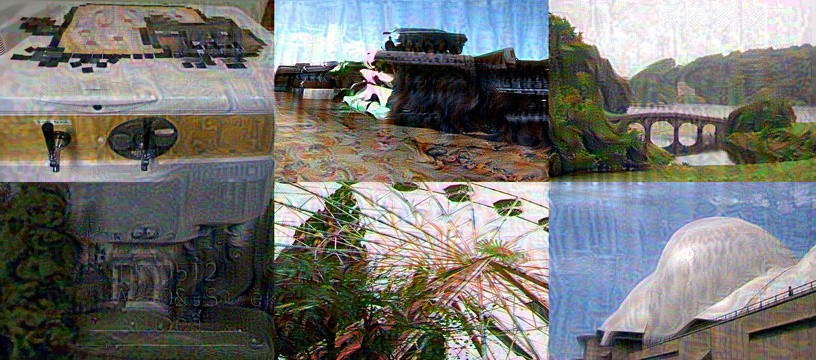} & \discimage{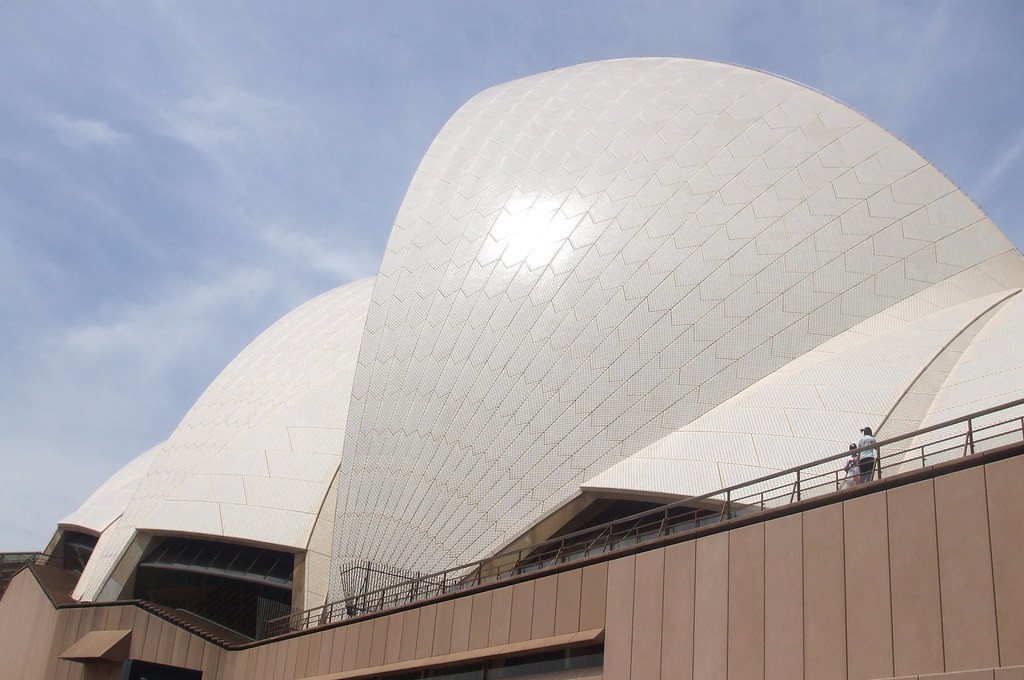} & \imcredsF{applez} \\    %
    \multicolumn{2}{c}{rank 9601} & 
    \multicolumn{2}{c}{rank 10070} \\
    \discimage{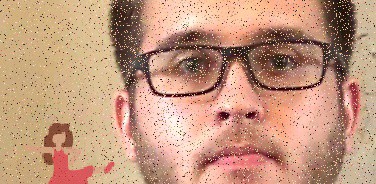} & \discimage{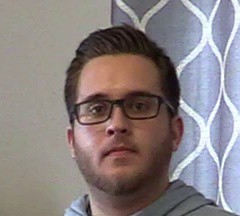} & \imcredsD{3158\_13} & %
    \discimage{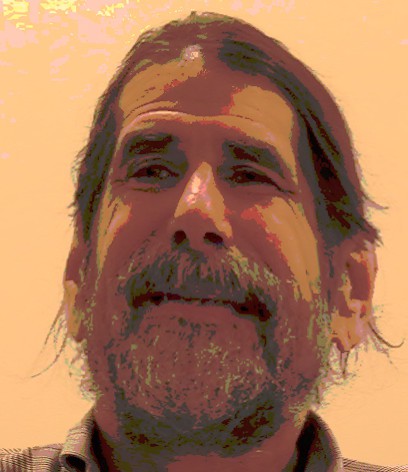} & \discimage{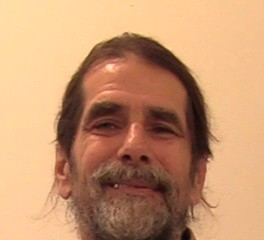} & \imcredsD{1859\_12} \\
    \end{tabular}
    \caption{
        Example false positive matches (pairs of query images and detected references images) from the submission \teamname{separate} (matching track). 
        The corresponding ranks of the query in the ranked list of all queries. 
        The smaller the ranking of a false positive is, the more it harms the evaluation metric.
    }
    \label{fig:exampleFPs}
\end{figure}

\paragraph{Manual vs. automatic transformations}

The image manipulations are either performed manually or via a series of carefully calibrated automatic transformations. Figure~\ref{fig:breakdownpersource} (right) compares the performance of the submissions depending on the type of transformations. It appears that the manual transformations are generally easier than the automatic ones. 

\paragraph{Analysis per manual editor}

The creation of the DISC21 involved 11 manual editors who produced 4040 query images. 
They were given instructions to make the transformations very difficult. 
However, Figure~\ref{fig:break_down_per_editor} shows that editor 44416353 made significantly more difficult transformations than the editor 45457694. 
Figure~\ref{fig:manual_edit_examples} shows example edits peformed by these two editors. 
The second editors did not perform strong geometric overlays, which make the task particularly difficult, as shown in Section~\ref{sec:break_down_per_transform}. 

In the following, the analysis focuses on automatic transformations, for which we have precise metadata on the types of transformations that are applied. 

\begin{figure}
    \centering
\floatbox[{\capbeside\thisfloatsetup{capbesideposition={left,top},capbesidewidth=7cm}}]{figure}[\FBwidth]
{
    \caption{
        Per-editor statistics for manual transformations/manipulations. 
        Each bar represents one editor, the width corresponds to how many query images the editor produced, and the height to the mAP for those queries in \teamname{lyakaap}'s submission for the descriptor track. 
        Editors are identified by the long numbers on the top of some bars.
    \label{fig:break_down_per_editor}   }
}
{
    \includegraphics[width=1.05\linewidth]{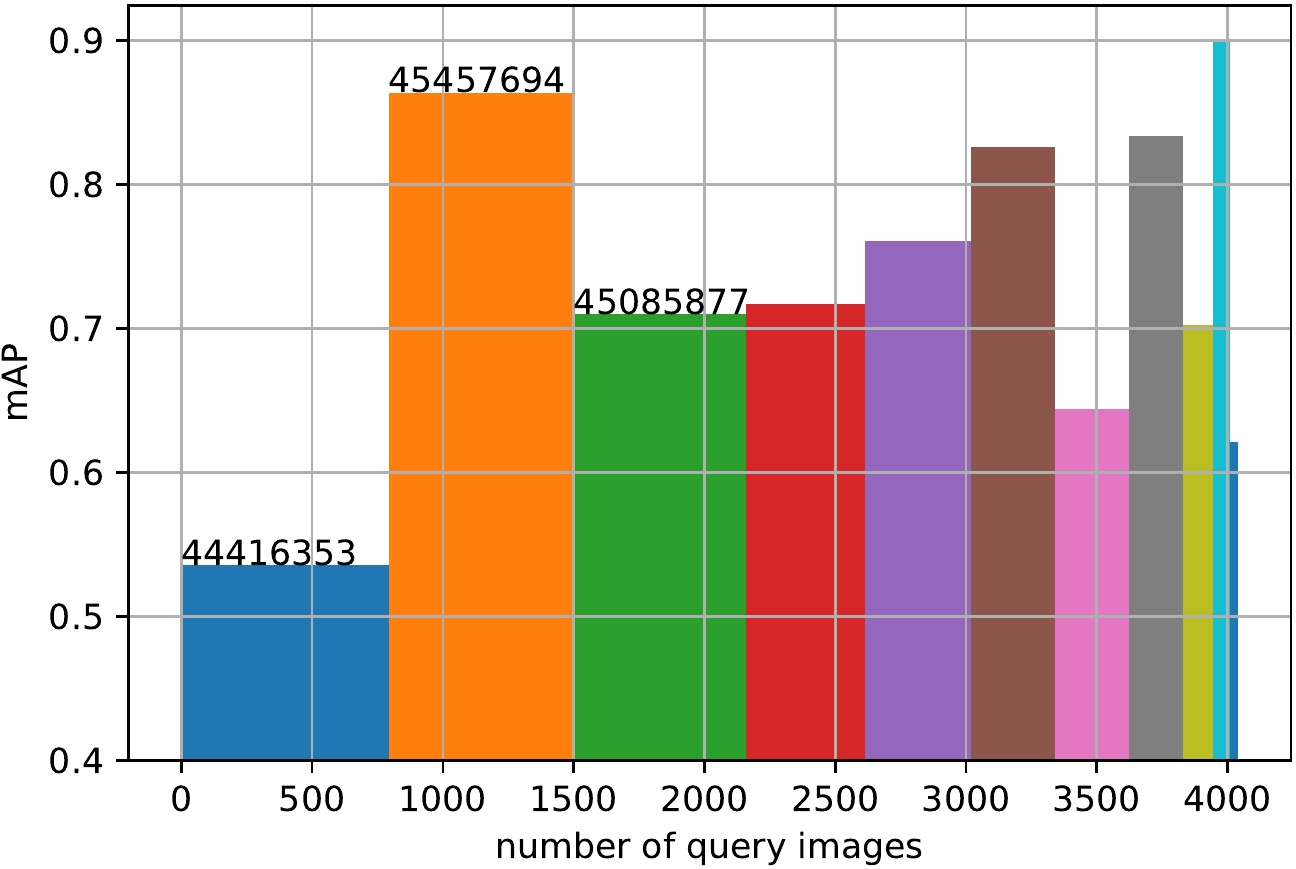}
}
\end{figure}

\begin{figure}
    \begin{tabular}{cc@{}c|cc@{}c}
    \multicolumn{2}{c}{Editor 44416353} &     \multicolumn{2}{c}{Editor 45457694} \\
    \discimage{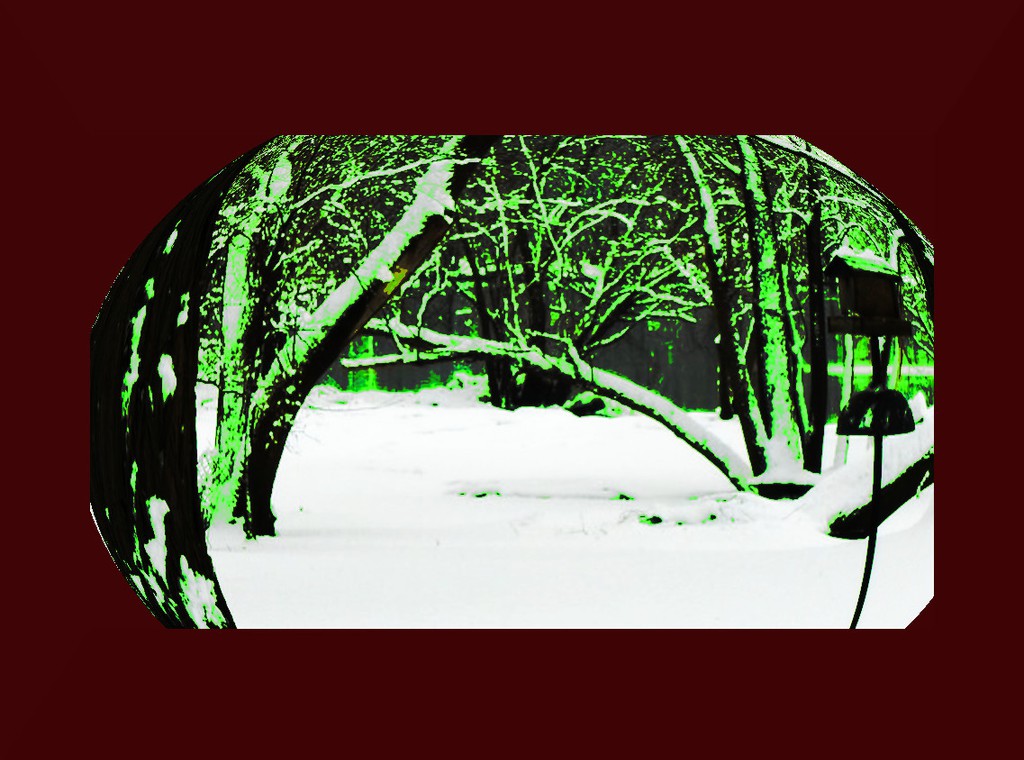} & \discimage{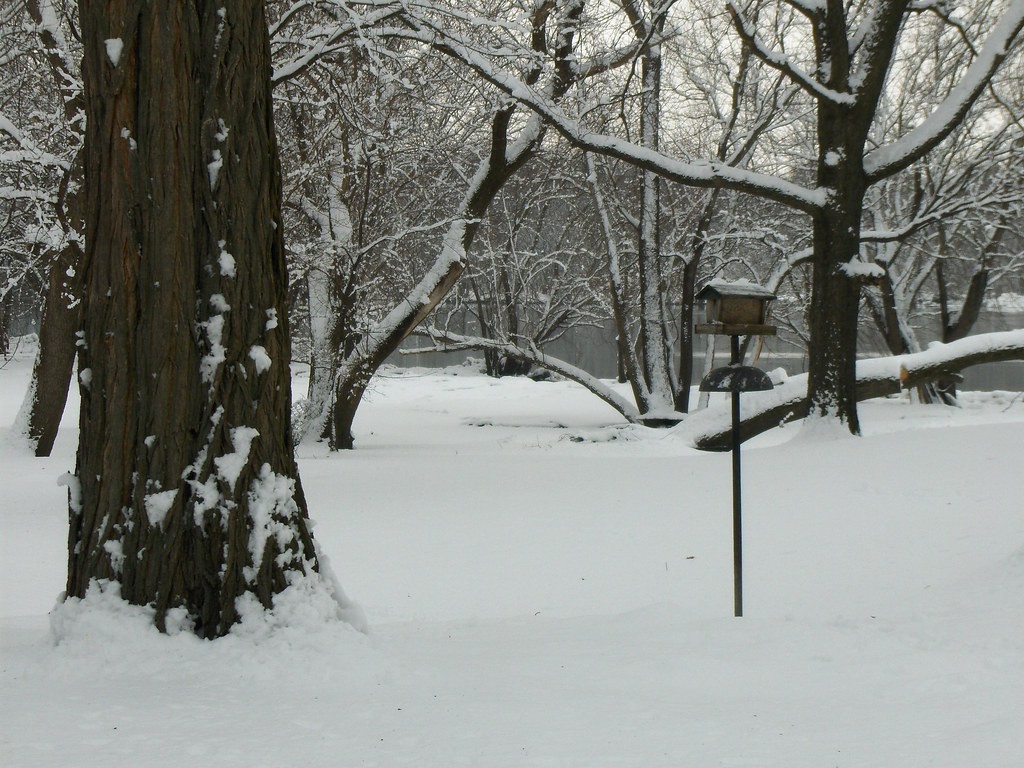} & \imcredsF{pixajen} & \discimage{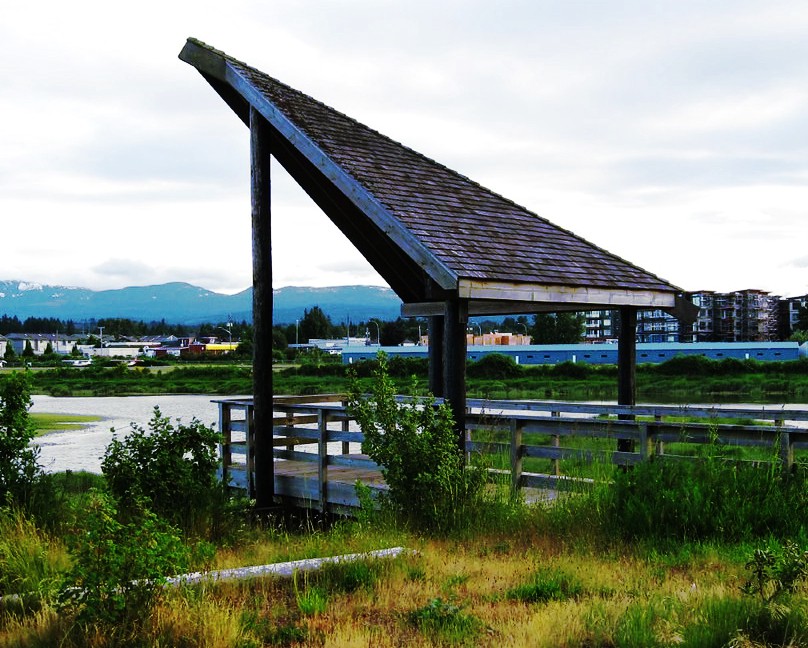} & \discimage{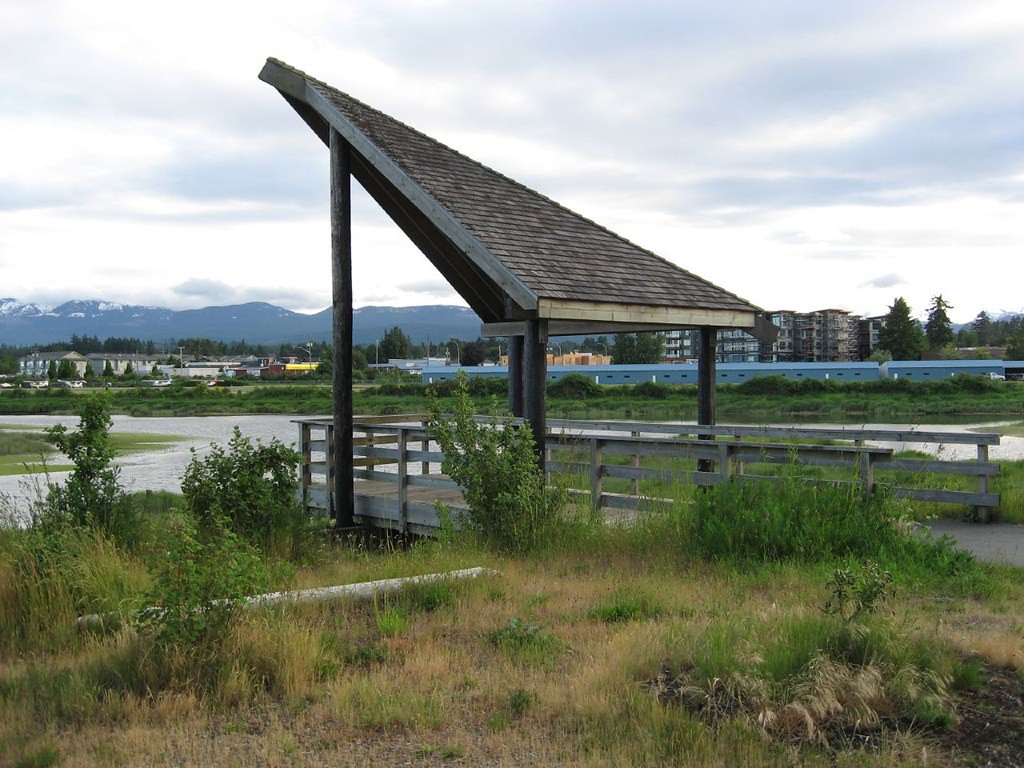} & \imcredsF{bchow} \\
    \discimage{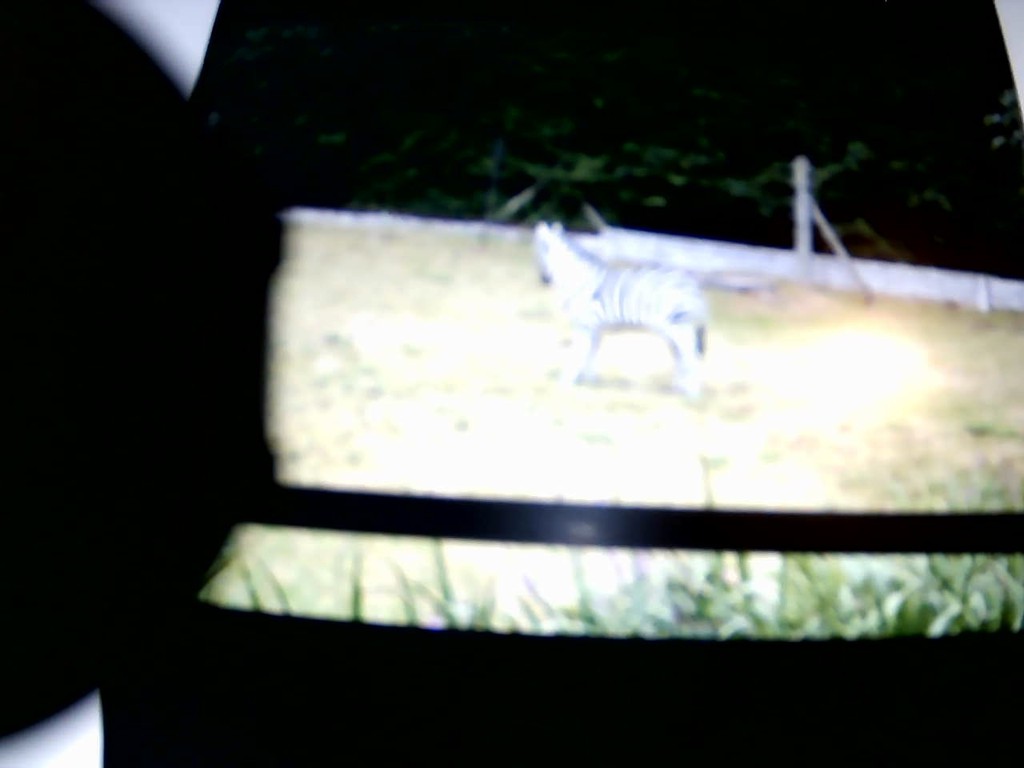} & \discimage{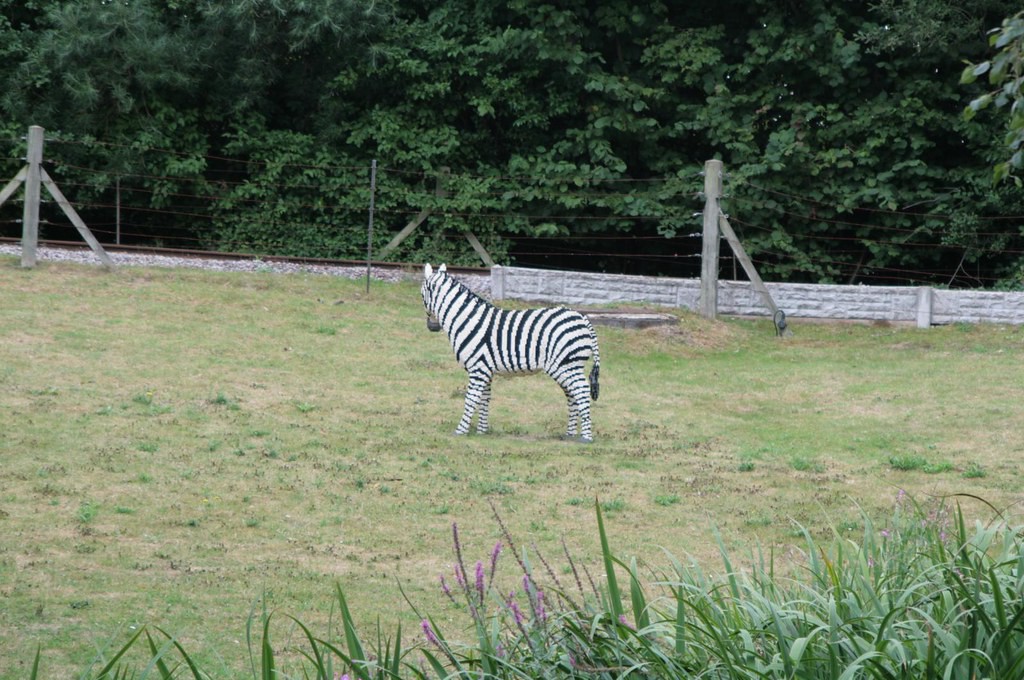} & \imcredsF{Karen Roe} & \discimage{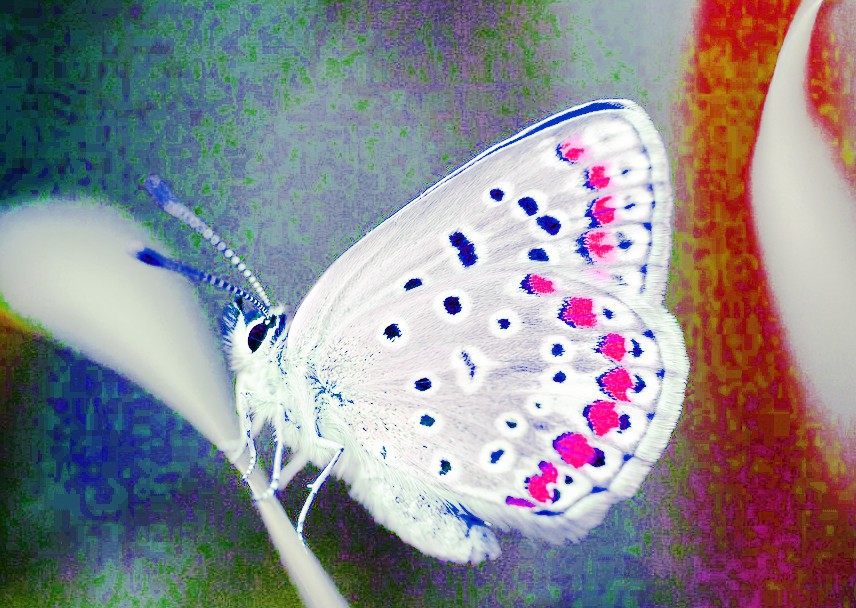} & \discimage{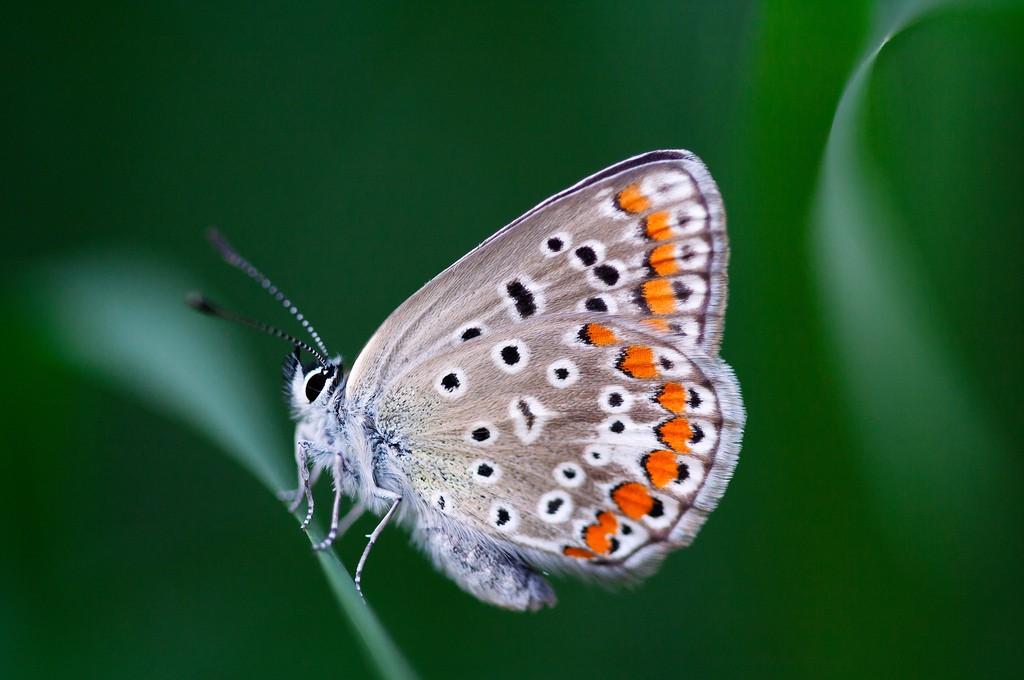} & \imcredsF{kaibara87} \\    
    \end{tabular}
    \caption{
        Examples of manual edits by the top-2 editors in Figure~\ref{fig:break_down_per_editor}.
    }
    \label{fig:manual_edit_examples}
\end{figure}

\subsection{Analysis per transformation} 
\label{sec:break_down_per_transform}

The automatically generated transformations were built by applying 2 to 6 transformations, in different steps, to all images. 
The random sampling of transformations was calibrated on the baseline matching methods at our disposal. 

\paragraph{Marginalized mAP measurements.}
Assessing the impact of each transformation type is not easy, because (1) there are only 132 query images that are produced with a \emph{single} transformation, (2) the intensity of most transformations depends on random parameters that are different between query images and (3) the impact on a retrieval measure like AP depends on the image content.
Therefore, there are not enough observations, \ie query images, to measure the impact of each transformation precisely. 
To mitigate this, we group the query images that have common transformation characteristics and compute the mAP within these groups. This marginalizes over the other transformations of the sequence.

\paragraph{Analysis per number of transformations.}

Since the focus of DISC21 is on difficult queries, the sampling was tuned to favor a large number of transformations, \eg 2340 query images are transformed with 4 steps, vs. 132 with a single one.
The transformations are grouped in classes (geometric, overlay, \etc) and at most one transformation per class is applied.  All the query images with 5 transformations include an adversarial attack step. 

Figure~\ref{fig:mAP_per_ntrans} shows the impact of the number transformations on mAP.
We observe the every extra transformation causes a larger performance drop than the previously added one, 
\ie the impact of a transformation is not independent of that of others:
for an image that is already hard to recognize, one additional transformation degrades the retrieval more than if it is applied to the original image.
Hence, the gap between methods is more marked for many transformations (more than 0.2 mAP) than on few (below 0.05 mAP).

\begin{figure}
\centering
    \includegraphics[width=0.48\linewidth]{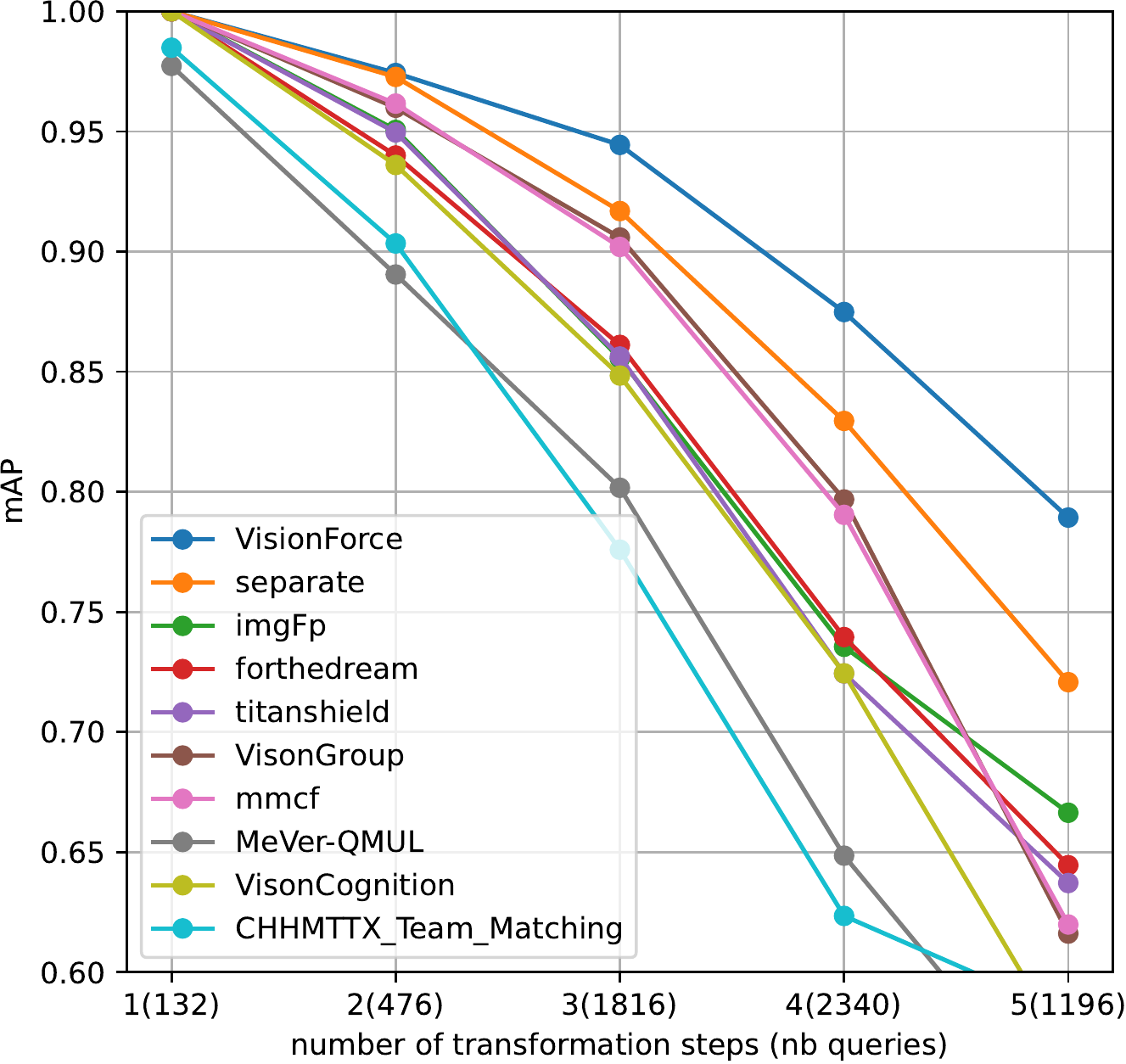}%
    \hfill
    \includegraphics[width=0.48\linewidth]{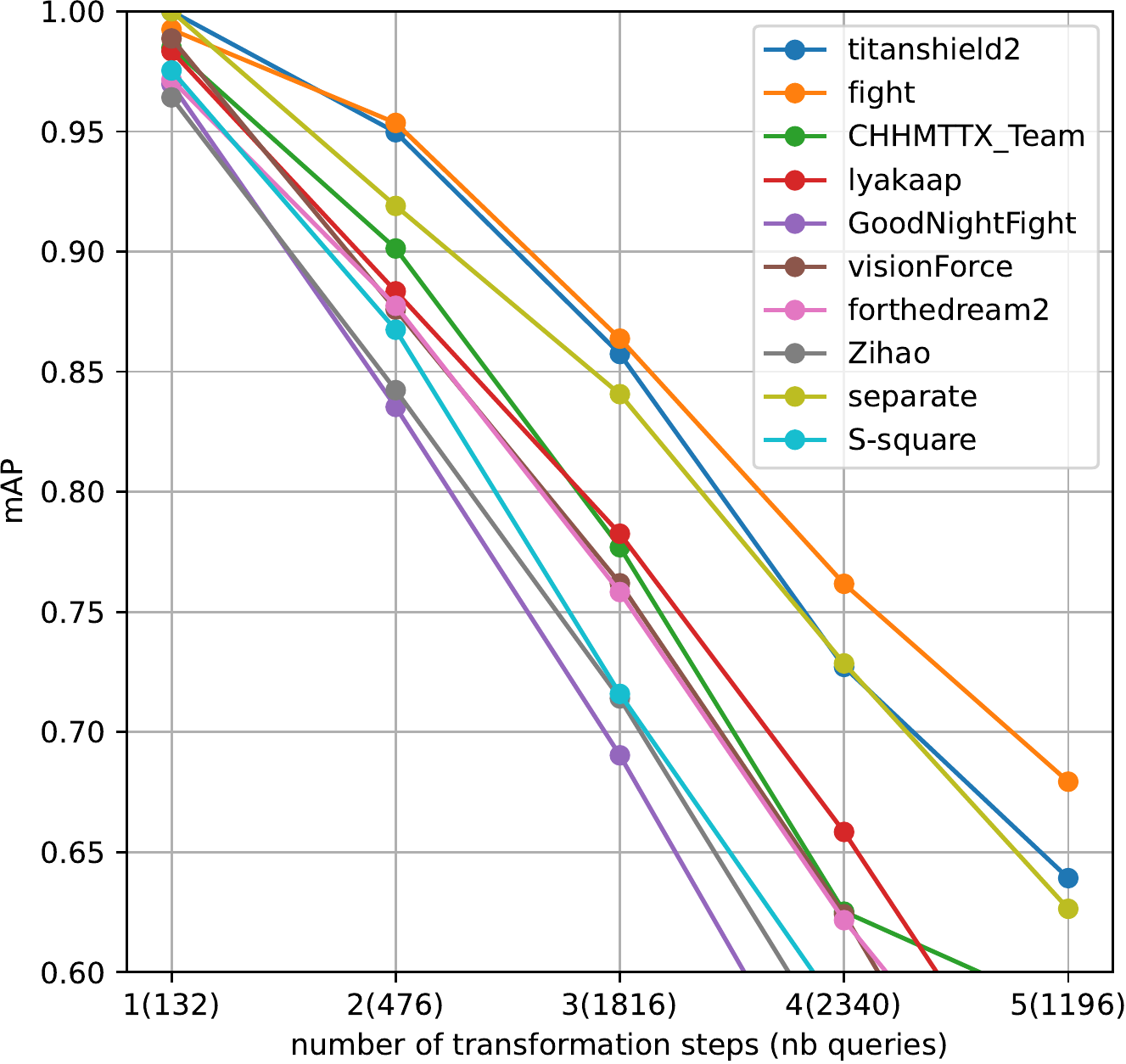}
    \caption{
        Performance (mAP) per number of transformations applied to create the query images, 
        in the matching track (left) and the descriptor track (right).
        The labels on the x-axis also indicate on how many images this mAP was computed.
    }
    \label{fig:mAP_per_ntrans}
\end{figure}

\subsection{Penalty analysis}

To further analyze the impact of each transformation, we use a \emph{penalty analysis} based on a simplistic model: 
for a given submission, we associate a fixed penalty to each of the 31 transformations, $(P_t)_{t=1..31}$. 
For a query image that undergoes transformation steps $(t_1, t_2,..., t_N)$, we model the resulting AP for that query as: 
\begin{equation}
    \mathrm{AP} = 1 - 
    \sum_{i=1}^N P_{t_i}
\end{equation}
This model is very rough, %
but its advantage is that the $(P_t)$ values can be estimated easily in the least squares sense from the per-query AP measurements.

Table~\ref{tab:tab_penalties_combined} shows that in general the hardest transformations are when the source image is inserted on top of an unrelated image.
This is hard to match with global descriptors in the descriptor track.

Geometric transformations are generally quite mild. 
Some submissions have particularly low performance on specific transformations, \eg \teamname{visionForce} has low performance on the vertical flip (vflip) transformation, and teams \teamname{CHHMTTX} and \teamname{GoodNightFight} struggle with rotations. 
It is likely that these geometric transformations were not included at training time for these submissions. 
The clip transformation seems to have a \emph{positive} impact on on the retrieval accuracy. 
This could be because the descriptor extraction at inference time often benefits from a stronger cropping than what is applied by default, See~\citep{touvron2019fixing}.

The penalties for the matching track are generally less severe than for the descriptor track. 
When using the full scale of matching techniques, some submissions like~\teamname{VisionForce} or \teamname{VisionGroup} become quite insensitive to geometric transformations where a large fraction of the image detail is removed.

In the following, we focus on a subset of the  transformations. 

\newcommand{\tcolhead}[1]{\rotatebox{90}{#1}~~}

\begin{table}
\rotatebox{270}{\scalebox{0.6}{
\begin{tabular}{rcr|rrrrrrrrrr|rrrrrrrrrrr}
&&& \multicolumn{10}{c|}{Matching track} & \multicolumn{10}{c}{Descriptor track} \\
  transformation    & \rotatebox{90}{type} & \tcolhead{nb of queries} &
   \tcolhead{VisionForce} & \tcolhead{separate} & \tcolhead{imgFp} & \tcolhead{forthedream} & \tcolhead{titanshield} & \tcolhead{VisonGroup} & \tcolhead{mmcf} & \tcolhead{MeVer-QMUL} & \tcolhead{VisonCognition} & \tcolhead{CHHMTTX} &
  \tcolhead{titanshield2} & \tcolhead{fight} & \tcolhead{CHHMTTX} & \tcolhead{lyakaap} & \tcolhead{GoodNightFight} & \tcolhead{visionForce} & \tcolhead{forthedream2} & \tcolhead{Zihao} & \tcolhead{separate} & \tcolhead{S-square} \\

\hline
change\_aspect\_ratio       & G &   474 		& \cellcolor[HTML]{8cff00}-0.02 & \cellcolor[HTML]{91ff00}0.01 & \cellcolor[HTML]{acff00}0.04 & \cellcolor[HTML]{8cff00}0.00 & \cellcolor[HTML]{8cff00}-0.02 & \cellcolor[HTML]{8cff00}-0.01 & \cellcolor[HTML]{8cff00}-0.02 & \cellcolor[HTML]{91ff00}0.01 & \cellcolor[HTML]{8cff00}-0.03 & \cellcolor[HTML]{8cff00}-0.02 & \cellcolor[HTML]{8cff00}-0.02			& \cellcolor[HTML]{8cff00}-0.01			& \cellcolor[HTML]{8cff00}-0.01			& \cellcolor[HTML]{8cff00}-0.01			& \cellcolor[HTML]{8cff00}-0.01			& \cellcolor[HTML]{8cff00}-0.00			& \cellcolor[HTML]{8cff00}-0.01			& \cellcolor[HTML]{97ff00}0.02			& \cellcolor[HTML]{8cff00}-0.00			& \cellcolor[HTML]{8cff00}-0.01\\
hflip                       & G &   498 		& \cellcolor[HTML]{8cff00}-0.05 & \cellcolor[HTML]{8cff00}-0.03 & \cellcolor[HTML]{d8ff00}0.09 & \cellcolor[HTML]{8cff00}-0.01 & \cellcolor[HTML]{8cff00}-0.03 & \cellcolor[HTML]{8cff00}-0.04 & \cellcolor[HTML]{8cff00}-0.04 & \cellcolor[HTML]{8cff00}-0.02 & \cellcolor[HTML]{8cff00}-0.03 & \cellcolor[HTML]{8cff00}-0.03 & \cellcolor[HTML]{8cff00}-0.03			& \cellcolor[HTML]{8cff00}-0.01			& \cellcolor[HTML]{8cff00}-0.03			& \cellcolor[HTML]{8cff00}-0.01			& \cellcolor[HTML]{8cff00}-0.02			& \cellcolor[HTML]{8cff00}-0.03			& \cellcolor[HTML]{8cff00}-0.04			& \cellcolor[HTML]{8cff00}-0.00			& \cellcolor[HTML]{8cff00}-0.04			& \cellcolor[HTML]{8cff00}-0.01\\
pad\_square                 & G &   494 		& \cellcolor[HTML]{8cff00}-0.01 & \cellcolor[HTML]{8cff00}-0.02 & \cellcolor[HTML]{8cff00}0.00 & \cellcolor[HTML]{8cff00}-0.01 & \cellcolor[HTML]{8cff00}-0.01 & \cellcolor[HTML]{8cff00}-0.02 & \cellcolor[HTML]{8cff00}-0.02 & \cellcolor[HTML]{97ff00}0.02 & \cellcolor[HTML]{8cff00}-0.02 & \cellcolor[HTML]{8cff00}0.00 & \cellcolor[HTML]{8cff00}-0.00			& \cellcolor[HTML]{8cff00}-0.01			& \cellcolor[HTML]{8cff00}0.00			& \cellcolor[HTML]{8cff00}-0.00			& \cellcolor[HTML]{91ff00}0.01			& \cellcolor[HTML]{8cff00}-0.02			& \cellcolor[HTML]{8cff00}-0.02			& \cellcolor[HTML]{9cff00}0.02			& \cellcolor[HTML]{8cff00}-0.02			& \cellcolor[HTML]{9cff00}0.02\\
perspective\_transform      & G &   450 		& \cellcolor[HTML]{acff00}0.04 & \cellcolor[HTML]{bdff00}0.06 & \cellcolor[HTML]{f8ff00}0.13 & \cellcolor[HTML]{c2ff00}0.07 & \cellcolor[HTML]{c2ff00}0.07 & \cellcolor[HTML]{c2ff00}0.07 & \cellcolor[HTML]{bdff00}0.06 & \cellcolor[HTML]{ddff00}0.10 & \cellcolor[HTML]{bdff00}0.06 & \cellcolor[HTML]{e8ff00}0.11 & \cellcolor[HTML]{c2ff00}0.07			& \cellcolor[HTML]{cdff00}0.08			& \cellcolor[HTML]{e8ff00}0.11			& \cellcolor[HTML]{d2ff00}0.09			& \cellcolor[HTML]{edff00}0.12			& \cellcolor[HTML]{c2ff00}0.06			& \cellcolor[HTML]{cdff00}0.08			& \cellcolor[HTML]{d8ff00}0.09			& \cellcolor[HTML]{cdff00}0.08			& \cellcolor[HTML]{cdff00}0.08\\
vflip                       & G &   474 		& \cellcolor[HTML]{8cff00}-0.01 & \cellcolor[HTML]{8cff00}-0.01 & \cellcolor[HTML]{ffdf00}0.18 & \cellcolor[HTML]{a7ff00}0.03 & \cellcolor[HTML]{8cff00}-0.02 & \cellcolor[HTML]{8cff00}-0.01 & \cellcolor[HTML]{8cff00}0.00 & \cellcolor[HTML]{b2ff00}0.05 & \cellcolor[HTML]{9cff00}0.02 & \cellcolor[HTML]{8cff00}-0.01 & \cellcolor[HTML]{8cff00}-0.02			& \cellcolor[HTML]{a7ff00}0.03			& \cellcolor[HTML]{8cff00}-0.01			& \cellcolor[HTML]{acff00}0.04			& \cellcolor[HTML]{8cff00}0.00			& \cellcolor[HTML]{ff4800}0.36			& \cellcolor[HTML]{a2ff00}0.03			& \cellcolor[HTML]{8cff00}-0.01			& \cellcolor[HTML]{8cff00}-0.01			& \cellcolor[HTML]{8cff00}-0.01\\
pixelization                & L &   677 		& \cellcolor[HTML]{91ff00}0.01 & \cellcolor[HTML]{97ff00}0.01 & \cellcolor[HTML]{8cff00}-0.00 & \cellcolor[HTML]{91ff00}0.01 & \cellcolor[HTML]{91ff00}0.01 & \cellcolor[HTML]{8cff00}-0.00 & \cellcolor[HTML]{8cff00}-0.00 & \cellcolor[HTML]{8cff00}-0.01 & \cellcolor[HTML]{8cff00}-0.02 & \cellcolor[HTML]{97ff00}0.01 & \cellcolor[HTML]{91ff00}0.01			& \cellcolor[HTML]{8cff00}0.00			& \cellcolor[HTML]{97ff00}0.01			& \cellcolor[HTML]{97ff00}0.01			& \cellcolor[HTML]{8cff00}0.00			& \cellcolor[HTML]{8cff00}-0.01			& \cellcolor[HTML]{97ff00}0.01			& \cellcolor[HTML]{97ff00}0.02			& \cellcolor[HTML]{a2ff00}0.03			& \cellcolor[HTML]{91ff00}0.01\\
overlay\_text               & O &  1737 		& \cellcolor[HTML]{9cff00}0.02 & \cellcolor[HTML]{91ff00}0.01 & \cellcolor[HTML]{acff00}0.04 & \cellcolor[HTML]{a2ff00}0.03 & \cellcolor[HTML]{9cff00}0.02 & \cellcolor[HTML]{9cff00}0.02 & \cellcolor[HTML]{a2ff00}0.03 & \cellcolor[HTML]{b7ff00}0.05 & \cellcolor[HTML]{a7ff00}0.04 & \cellcolor[HTML]{b2ff00}0.05 & \cellcolor[HTML]{97ff00}0.02			& \cellcolor[HTML]{acff00}0.04			& \cellcolor[HTML]{b2ff00}0.05			& \cellcolor[HTML]{97ff00}0.02			& \cellcolor[HTML]{b2ff00}0.05			& \cellcolor[HTML]{b7ff00}0.05			& \cellcolor[HTML]{acff00}0.04			& \cellcolor[HTML]{bdff00}0.06			& \cellcolor[HTML]{9cff00}0.02			& \cellcolor[HTML]{b7ff00}0.05\\
apply\_pil\_filter          & L &   631 		& \cellcolor[HTML]{a2ff00}0.03 & \cellcolor[HTML]{b2ff00}0.04 & \cellcolor[HTML]{bdff00}0.06 & \cellcolor[HTML]{acff00}0.04 & \cellcolor[HTML]{bdff00}0.06 & \cellcolor[HTML]{a2ff00}0.03 & \cellcolor[HTML]{9cff00}0.02 & \cellcolor[HTML]{97ff00}0.02 & \cellcolor[HTML]{9cff00}0.02 & \cellcolor[HTML]{acff00}0.04 & \cellcolor[HTML]{bdff00}0.06			& \cellcolor[HTML]{acff00}0.04			& \cellcolor[HTML]{acff00}0.04			& \cellcolor[HTML]{91ff00}0.01			& \cellcolor[HTML]{a2ff00}0.03			& \cellcolor[HTML]{acff00}0.04			& \cellcolor[HTML]{acff00}0.04			& \cellcolor[HTML]{a7ff00}0.03			& \cellcolor[HTML]{a7ff00}0.04			& \cellcolor[HTML]{8cff00}-0.01\\
encoding\_quality           & L &   644 		& \cellcolor[HTML]{a7ff00}0.04 & \cellcolor[HTML]{97ff00}0.01 & \cellcolor[HTML]{a7ff00}0.03 & \cellcolor[HTML]{9cff00}0.02 & \cellcolor[HTML]{97ff00}0.01 & \cellcolor[HTML]{91ff00}0.01 & \cellcolor[HTML]{97ff00}0.01 & \cellcolor[HTML]{97ff00}0.01 & \cellcolor[HTML]{8cff00}0.00 & \cellcolor[HTML]{acff00}0.04 & \cellcolor[HTML]{97ff00}0.01			& \cellcolor[HTML]{8cff00}0.00			& \cellcolor[HTML]{a7ff00}0.04			& \cellcolor[HTML]{8cff00}-0.01			& \cellcolor[HTML]{a2ff00}0.03			& \cellcolor[HTML]{a2ff00}0.03			& \cellcolor[HTML]{a7ff00}0.03			& \cellcolor[HTML]{9cff00}0.02			& \cellcolor[HTML]{8cff00}0.00			& \cellcolor[HTML]{8cff00}-0.01\\
apply\_ig\_filter           & P &  1066 		& \cellcolor[HTML]{91ff00}0.01 & \cellcolor[HTML]{97ff00}0.02 & \cellcolor[HTML]{97ff00}0.01 & \cellcolor[HTML]{a7ff00}0.03 & \cellcolor[HTML]{bdff00}0.06 & \cellcolor[HTML]{a7ff00}0.04 & \cellcolor[HTML]{a7ff00}0.04 & \cellcolor[HTML]{97ff00}0.02 & \cellcolor[HTML]{b2ff00}0.05 & \cellcolor[HTML]{a7ff00}0.03 & \cellcolor[HTML]{bdff00}0.06			& \cellcolor[HTML]{97ff00}0.01			& \cellcolor[HTML]{a2ff00}0.03			& \cellcolor[HTML]{a7ff00}0.04			& \cellcolor[HTML]{a2ff00}0.03			& \cellcolor[HTML]{a2ff00}0.03			& \cellcolor[HTML]{a7ff00}0.03			& \cellcolor[HTML]{97ff00}0.01			& \cellcolor[HTML]{a2ff00}0.03			& \cellcolor[HTML]{a7ff00}0.03\\
overlay\_emoji              & O &  1753 		& \cellcolor[HTML]{a2ff00}0.03 & \cellcolor[HTML]{9cff00}0.02 & \cellcolor[HTML]{c2ff00}0.07 & \cellcolor[HTML]{b7ff00}0.05 & \cellcolor[HTML]{b2ff00}0.05 & \cellcolor[HTML]{b7ff00}0.05 & \cellcolor[HTML]{bdff00}0.06 & \cellcolor[HTML]{bdff00}0.06 & \cellcolor[HTML]{c7ff00}0.07 & \cellcolor[HTML]{c7ff00}0.07 & \cellcolor[HTML]{b2ff00}0.05			& \cellcolor[HTML]{b7ff00}0.06			& \cellcolor[HTML]{c2ff00}0.07			& \cellcolor[HTML]{bdff00}0.06			& \cellcolor[HTML]{bdff00}0.06			& \cellcolor[HTML]{bdff00}0.06			& \cellcolor[HTML]{d8ff00}0.09			& \cellcolor[HTML]{b7ff00}0.05			& \cellcolor[HTML]{a7ff00}0.04			& \cellcolor[HTML]{cdff00}0.08\\
grayscale                   & P &  1018 		& \cellcolor[HTML]{91ff00}0.01 & \cellcolor[HTML]{a2ff00}0.03 & \cellcolor[HTML]{b2ff00}0.05 & \cellcolor[HTML]{c7ff00}0.07 & \cellcolor[HTML]{cdff00}0.08 & \cellcolor[HTML]{c7ff00}0.07 & \cellcolor[HTML]{c7ff00}0.07 & \cellcolor[HTML]{c7ff00}0.07 & \cellcolor[HTML]{cdff00}0.08 & \cellcolor[HTML]{bdff00}0.06 & \cellcolor[HTML]{cdff00}0.08			& \cellcolor[HTML]{a2ff00}0.03			& \cellcolor[HTML]{b7ff00}0.06			& \cellcolor[HTML]{b2ff00}0.05			& \cellcolor[HTML]{bdff00}0.06			& \cellcolor[HTML]{b7ff00}0.06			& \cellcolor[HTML]{c2ff00}0.07			& \cellcolor[HTML]{b2ff00}0.05			& \cellcolor[HTML]{c2ff00}0.06			& \cellcolor[HTML]{c7ff00}0.07\\
rotate                      & G &   417 		& \cellcolor[HTML]{c7ff00}0.07 & \cellcolor[HTML]{d2ff00}0.08 & \cellcolor[HTML]{d2ff00}0.08 & \cellcolor[HTML]{d2ff00}0.08 & \cellcolor[HTML]{b7ff00}0.05 & \cellcolor[HTML]{ddff00}0.10 & \cellcolor[HTML]{e8ff00}0.11 & \cellcolor[HTML]{fffa00}0.14 & \cellcolor[HTML]{e8ff00}0.11 & \cellcolor[HTML]{ff6800}0.32 & \cellcolor[HTML]{b2ff00}0.05			& \cellcolor[HTML]{c7ff00}0.07			& \cellcolor[HTML]{ff6800}0.32			& \cellcolor[HTML]{e3ff00}0.11			& \cellcolor[HTML]{ff6300}0.33			& \cellcolor[HTML]{ffdf00}0.18			& \cellcolor[HTML]{f8ff00}0.13			& \cellcolor[HTML]{f8ff00}0.13			& \cellcolor[HTML]{b7ff00}0.06			& \cellcolor[HTML]{fffa00}0.15\\
saturation                  & P &  1021 		& \cellcolor[HTML]{9cff00}0.02 & \cellcolor[HTML]{a2ff00}0.03 & \cellcolor[HTML]{a7ff00}0.03 & \cellcolor[HTML]{b7ff00}0.05 & \cellcolor[HTML]{a2ff00}0.03 & \cellcolor[HTML]{a2ff00}0.03 & \cellcolor[HTML]{a7ff00}0.04 & \cellcolor[HTML]{acff00}0.04 & \cellcolor[HTML]{b2ff00}0.04 & \cellcolor[HTML]{b7ff00}0.06 & \cellcolor[HTML]{a2ff00}0.03			& \cellcolor[HTML]{9cff00}0.02			& \cellcolor[HTML]{b7ff00}0.06			& \cellcolor[HTML]{b7ff00}0.05			& \cellcolor[HTML]{b7ff00}0.06			& \cellcolor[HTML]{cdff00}0.08			& \cellcolor[HTML]{c2ff00}0.07			& \cellcolor[HTML]{acff00}0.04			& \cellcolor[HTML]{b7ff00}0.06			& \cellcolor[HTML]{a2ff00}0.03\\
shuffle\_pixels             & L &   643 		& \cellcolor[HTML]{acff00}0.04 & \cellcolor[HTML]{bdff00}0.06 & \cellcolor[HTML]{a7ff00}0.03 & \cellcolor[HTML]{bdff00}0.06 & \cellcolor[HTML]{bdff00}0.06 & \cellcolor[HTML]{bdff00}0.06 & \cellcolor[HTML]{c2ff00}0.06 & \cellcolor[HTML]{b7ff00}0.05 & \cellcolor[HTML]{bdff00}0.06 & \cellcolor[HTML]{d2ff00}0.09 & \cellcolor[HTML]{bdff00}0.06			& \cellcolor[HTML]{a7ff00}0.03			& \cellcolor[HTML]{d2ff00}0.09			& \cellcolor[HTML]{acff00}0.04			& \cellcolor[HTML]{cdff00}0.08			& \cellcolor[HTML]{9cff00}0.02			& \cellcolor[HTML]{bdff00}0.06			& \cellcolor[HTML]{b7ff00}0.06			& \cellcolor[HTML]{b7ff00}0.05			& \cellcolor[HTML]{9cff00}0.02\\
clip\_image\_size           & G &  7161 		& \cellcolor[HTML]{8cff00}-0.10 & \cellcolor[HTML]{8cff00}-0.13 & \cellcolor[HTML]{8cff00}-0.16 & \cellcolor[HTML]{8cff00}-0.17 & \cellcolor[HTML]{8cff00}-0.17 & \cellcolor[HTML]{8cff00}-0.14 & \cellcolor[HTML]{8cff00}-0.15 & \cellcolor[HTML]{8cff00}-0.16 & \cellcolor[HTML]{8cff00}-0.16 & \cellcolor[HTML]{8cff00}-0.16 & \cellcolor[HTML]{8cff00}-0.17			& \cellcolor[HTML]{8cff00}-0.13			& \cellcolor[HTML]{8cff00}-0.16			& \cellcolor[HTML]{8cff00}-0.12			& \cellcolor[HTML]{8cff00}-0.13			& \cellcolor[HTML]{8cff00}-0.14			& \cellcolor[HTML]{8cff00}-0.14			& \cellcolor[HTML]{8cff00}-0.12			& \cellcolor[HTML]{8cff00}-0.12			& \cellcolor[HTML]{8cff00}-0.12\\
random\_noise               & L &   621 		& \cellcolor[HTML]{bdff00}0.06 & \cellcolor[HTML]{e8ff00}0.11 & \cellcolor[HTML]{c2ff00}0.07 & \cellcolor[HTML]{f8ff00}0.13 & \cellcolor[HTML]{ddff00}0.10 & \cellcolor[HTML]{ddff00}0.10 & \cellcolor[HTML]{d8ff00}0.10 & \cellcolor[HTML]{b7ff00}0.06 & \cellcolor[HTML]{d2ff00}0.09 & \cellcolor[HTML]{d8ff00}0.10 & \cellcolor[HTML]{ddff00}0.10			& \cellcolor[HTML]{bdff00}0.06			& \cellcolor[HTML]{d8ff00}0.10			& \cellcolor[HTML]{d2ff00}0.09			& \cellcolor[HTML]{c7ff00}0.08			& \cellcolor[HTML]{d8ff00}0.09			& \cellcolor[HTML]{fff500}0.15			& \cellcolor[HTML]{c2ff00}0.07			& \cellcolor[HTML]{d8ff00}0.10			& \cellcolor[HTML]{c2ff00}0.07\\
brightness                  & P &  1009 		& \cellcolor[HTML]{b7ff00}0.05 & \cellcolor[HTML]{d8ff00}0.09 & \cellcolor[HTML]{c7ff00}0.07 & \cellcolor[HTML]{ddff00}0.10 & \cellcolor[HTML]{c7ff00}0.07 & \cellcolor[HTML]{c7ff00}0.07 & \cellcolor[HTML]{c7ff00}0.07 & \cellcolor[HTML]{c7ff00}0.07 & \cellcolor[HTML]{c7ff00}0.07 & \cellcolor[HTML]{e3ff00}0.10 & \cellcolor[HTML]{c7ff00}0.07			& \cellcolor[HTML]{b7ff00}0.05			& \cellcolor[HTML]{ddff00}0.10			& \cellcolor[HTML]{d8ff00}0.10			& \cellcolor[HTML]{c7ff00}0.07			& \cellcolor[HTML]{c7ff00}0.07			& \cellcolor[HTML]{ddff00}0.10			& \cellcolor[HTML]{b7ff00}0.06			& \cellcolor[HTML]{d8ff00}0.10			& \cellcolor[HTML]{c2ff00}0.07\\
legofy                      & L &   633 		& \cellcolor[HTML]{bdff00}0.06 & \cellcolor[HTML]{ddff00}0.10 & \cellcolor[HTML]{acff00}0.04 & \cellcolor[HTML]{e8ff00}0.11 & \cellcolor[HTML]{bdff00}0.06 & \cellcolor[HTML]{e3ff00}0.11 & \cellcolor[HTML]{e8ff00}0.11 & \cellcolor[HTML]{cdff00}0.08 & \cellcolor[HTML]{e8ff00}0.12 & \cellcolor[HTML]{c7ff00}0.07 & \cellcolor[HTML]{bdff00}0.06			& \cellcolor[HTML]{a7ff00}0.04			& \cellcolor[HTML]{c2ff00}0.07			& \cellcolor[HTML]{cdff00}0.08			& \cellcolor[HTML]{c2ff00}0.06			& \cellcolor[HTML]{cdff00}0.08			& \cellcolor[HTML]{feff00}0.14			& \cellcolor[HTML]{edff00}0.12			& \cellcolor[HTML]{c7ff00}0.07			& \cellcolor[HTML]{b2ff00}0.05\\
apply\_ar\_effect           & O &  1049 		& \cellcolor[HTML]{c7ff00}0.07 & \cellcolor[HTML]{d2ff00}0.09 & \cellcolor[HTML]{e3ff00}0.11 & \cellcolor[HTML]{f3ff00}0.12 & \cellcolor[HTML]{ffd400}0.19 & \cellcolor[HTML]{fffa00}0.14 & \cellcolor[HTML]{feff00}0.14 & \cellcolor[HTML]{fffa00}0.14 & \cellcolor[HTML]{fff500}0.15 & \cellcolor[HTML]{edff00}0.12 & \cellcolor[HTML]{ffd400}0.19			& \cellcolor[HTML]{d2ff00}0.09			& \cellcolor[HTML]{edff00}0.12			& \cellcolor[HTML]{e3ff00}0.10			& \cellcolor[HTML]{e3ff00}0.11			& \cellcolor[HTML]{fffa00}0.15			& \cellcolor[HTML]{f8ff00}0.13			& \cellcolor[HTML]{feff00}0.14			& \cellcolor[HTML]{e8ff00}0.11			& \cellcolor[HTML]{ffe400}0.17\\
convert\_color              & P &  1871 		& \cellcolor[HTML]{fffa00}0.15 & \cellcolor[HTML]{feff00}0.14 & \cellcolor[HTML]{ffe400}0.17 & \cellcolor[HTML]{ffea00}0.17 & \cellcolor[HTML]{ffc400}0.21 & \cellcolor[HTML]{ffea00}0.16 & \cellcolor[HTML]{ffef00}0.16 & \cellcolor[HTML]{fffa00}0.14 & \cellcolor[HTML]{ffea00}0.16 & \cellcolor[HTML]{fffa00}0.14 & \cellcolor[HTML]{ffc900}0.21			& \cellcolor[HTML]{f8ff00}0.13			& \cellcolor[HTML]{fffa00}0.14			& \cellcolor[HTML]{fffa00}0.15			& \cellcolor[HTML]{e8ff00}0.11			& \cellcolor[HTML]{ffc400}0.21			& \cellcolor[HTML]{ffda00}0.18			& \cellcolor[HTML]{e8ff00}0.11			& \cellcolor[HTML]{fff500}0.15			& \cellcolor[HTML]{f8ff00}0.13\\
overlay\_stripes            & O &  1735 		& \cellcolor[HTML]{ffea00}0.16 & \cellcolor[HTML]{ffef00}0.15 & \cellcolor[HTML]{ff8300}0.29 & \cellcolor[HTML]{ff9e00}0.26 & \cellcolor[HTML]{ffda00}0.19 & \cellcolor[HTML]{ffd400}0.19 & \cellcolor[HTML]{ffbf00}0.22 & \cellcolor[HTML]{ffb900}0.22 & \cellcolor[HTML]{ffa900}0.24 & \cellcolor[HTML]{ffc400}0.21 & \cellcolor[HTML]{ffda00}0.18			& \cellcolor[HTML]{ffcf00}0.20			& \cellcolor[HTML]{ffc400}0.21			& \cellcolor[HTML]{ffbf00}0.22			& \cellcolor[HTML]{ffea00}0.17			& \cellcolor[HTML]{ff9900}0.26			& \cellcolor[HTML]{ff7800}0.30			& \cellcolor[HTML]{ddff00}0.10			& \cellcolor[HTML]{f8ff00}0.14			& \cellcolor[HTML]{ffa400}0.25\\
blur                        & L &   648 		& \cellcolor[HTML]{f8ff00}0.13 & \cellcolor[HTML]{ff7300}0.31 & \cellcolor[HTML]{ff8e00}0.28 & \cellcolor[HTML]{ffb900}0.22 & \cellcolor[HTML]{ffda00}0.18 & \cellcolor[HTML]{ffd400}0.19 & \cellcolor[HTML]{ffcf00}0.20 & \cellcolor[HTML]{ffb400}0.23 & \cellcolor[HTML]{ffdf00}0.18 & \cellcolor[HTML]{ffea00}0.17 & \cellcolor[HTML]{ffda00}0.18			& \cellcolor[HTML]{ffd400}0.19			& \cellcolor[HTML]{ffea00}0.17			& \cellcolor[HTML]{ffd400}0.19			& \cellcolor[HTML]{f3ff00}0.12			& \cellcolor[HTML]{ffa900}0.24			& \cellcolor[HTML]{ffae00}0.23			& \cellcolor[HTML]{ff6e00}0.32			& \cellcolor[HTML]{ffb400}0.23			& \cellcolor[HTML]{ffe400}0.17\\
adversarial\_attack         & A &  1201 		& \cellcolor[HTML]{8cff00}-0.04 & \cellcolor[HTML]{8cff00}-0.01 & \cellcolor[HTML]{8cff00}-0.08 & \cellcolor[HTML]{8cff00}-0.05 & \cellcolor[HTML]{8cff00}-0.09 & \cellcolor[HTML]{b2ff00}0.04 & \cellcolor[HTML]{acff00}0.04 & \cellcolor[HTML]{8cff00}-0.01 & \cellcolor[HTML]{9cff00}0.02 & \cellcolor[HTML]{8cff00}-0.08 & \cellcolor[HTML]{8cff00}-0.08			& \cellcolor[HTML]{8cff00}-0.03			& \cellcolor[HTML]{8cff00}-0.08			& \cellcolor[HTML]{97ff00}0.02			& \cellcolor[HTML]{8cff00}-0.07			& \cellcolor[HTML]{8cff00}-0.03			& \cellcolor[HTML]{8cff00}-0.04			& \cellcolor[HTML]{d8ff00}0.10			& \cellcolor[HTML]{8cff00}-0.02			& \cellcolor[HTML]{8cff00}-0.01\\
overlay\_onto\_screenshot   & I &   438 		& \cellcolor[HTML]{edff00}0.12 & \cellcolor[HTML]{ffcf00}0.20 & \cellcolor[HTML]{ffbf00}0.21 & \cellcolor[HTML]{ffcf00}0.20 & \cellcolor[HTML]{ff3700}0.38 & \cellcolor[HTML]{fff500}0.15 & \cellcolor[HTML]{fff500}0.15 & \cellcolor[HTML]{ff0028}0.53 & \cellcolor[HTML]{ffbf00}0.22 & \cellcolor[HTML]{ff0028}0.52 & \cellcolor[HTML]{ff3700}0.38			& \cellcolor[HTML]{ffa400}0.25			& \cellcolor[HTML]{ff0028}0.52			& \cellcolor[HTML]{ff4d00}0.35			& \cellcolor[HTML]{ff0028}0.73			& \cellcolor[HTML]{ff1200}0.42			& \cellcolor[HTML]{ff6e00}0.32			& \cellcolor[HTML]{ff0013}0.47			& \cellcolor[HTML]{ff8e00}0.28			& \cellcolor[HTML]{ff6300}0.33\\
crop                        & G &   468 		& \cellcolor[HTML]{ffa900}0.24 & \cellcolor[HTML]{ff9900}0.26 & \cellcolor[HTML]{ff9900}0.26 & \cellcolor[HTML]{ff0003}0.45 & \cellcolor[HTML]{ff4800}0.36 & \cellcolor[HTML]{ff6300}0.33 & \cellcolor[HTML]{ff5800}0.34 & \cellcolor[HTML]{ff1c00}0.41 & \cellcolor[HTML]{ff5800}0.34 & \cellcolor[HTML]{ff0028}0.63 & \cellcolor[HTML]{ff4d00}0.35			& \cellcolor[HTML]{ff0c00}0.43			& \cellcolor[HTML]{ff0028}0.63			& \cellcolor[HTML]{ff1200}0.42			& \cellcolor[HTML]{ff0028}0.64			& \cellcolor[HTML]{ff4200}0.37			& \cellcolor[HTML]{ff0028}0.64			& \cellcolor[HTML]{ff0028}0.57			& \cellcolor[HTML]{ff0028}0.49			& \cellcolor[HTML]{ff0028}0.58\\
overlay\_onto\_image        & I &   470 		& \cellcolor[HTML]{c2ff00}0.07 & \cellcolor[HTML]{ffe400}0.17 & \cellcolor[HTML]{ffb900}0.22 & \cellcolor[HTML]{ffea00}0.16 & \cellcolor[HTML]{ff5d00}0.33 & \cellcolor[HTML]{ffdf00}0.18 & \cellcolor[HTML]{fffa00}0.14 & \cellcolor[HTML]{ff0028}0.53 & \cellcolor[HTML]{ffcf00}0.20 & \cellcolor[HTML]{ff0028}0.54 & \cellcolor[HTML]{ff6300}0.33			& \cellcolor[HTML]{ff6800}0.32			& \cellcolor[HTML]{ff0028}0.54			& \cellcolor[HTML]{ff0c00}0.43			& \cellcolor[HTML]{ff0028}0.88			& \cellcolor[HTML]{ffc900}0.20			& \cellcolor[HTML]{ffdf00}0.18			& \cellcolor[HTML]{ff0028}0.81			& \cellcolor[HTML]{ffc900}0.20			& \cellcolor[HTML]{ff0028}0.54\\
overlay\_blurred\_mask      & I &   479 		& \cellcolor[HTML]{cdff00}0.08 & \cellcolor[HTML]{ffcf00}0.20 & \cellcolor[HTML]{ff3d00}0.38 & \cellcolor[HTML]{ffda00}0.19 & \cellcolor[HTML]{ff0700}0.43 & \cellcolor[HTML]{ffa900}0.24 & \cellcolor[HTML]{ffb400}0.23 & \cellcolor[HTML]{ff0028}0.51 & \cellcolor[HTML]{ff7800}0.30 & \cellcolor[HTML]{ff0028}0.50 & \cellcolor[HTML]{ff0c00}0.43			& \cellcolor[HTML]{ff1c00}0.41			& \cellcolor[HTML]{ff0028}0.50			& \cellcolor[HTML]{ff0c00}0.43			& \cellcolor[HTML]{ff0028}0.76			& \cellcolor[HTML]{ff0023}0.49			& \cellcolor[HTML]{ffa900}0.24			& \cellcolor[HTML]{ff0028}0.79			& \cellcolor[HTML]{ff9300}0.27			& \cellcolor[HTML]{ff0028}0.74\\
collage                     & I &   454 		& \cellcolor[HTML]{a7ff00}0.04 & \cellcolor[HTML]{b7ff00}0.05 & \cellcolor[HTML]{cdff00}0.08 & \cellcolor[HTML]{ff3200}0.39 & \cellcolor[HTML]{ff0023}0.49 & \cellcolor[HTML]{e3ff00}0.11 & \cellcolor[HTML]{ddff00}0.10 & \cellcolor[HTML]{ff0100}0.45 & \cellcolor[HTML]{ff0028}0.56 & \cellcolor[HTML]{ff0c00}0.43 & \cellcolor[HTML]{ff0023}0.49			& \cellcolor[HTML]{ff9300}0.27			& \cellcolor[HTML]{ff1200}0.42			& \cellcolor[HTML]{ff0028}0.65			& \cellcolor[HTML]{ff0028}0.76			& \cellcolor[HTML]{ff1200}0.42			& \cellcolor[HTML]{ff0028}0.62			& \cellcolor[HTML]{ff0028}0.72			& \cellcolor[HTML]{ff0028}0.63			& \cellcolor[HTML]{ff0028}0.71\\
\end{tabular}

}}
    \caption{
Relative penalties for all transformations.
The transformations are classified into 5 types: G=geometric, L=local pixel transformation, O=source image is partially occluded, P=photometric transformation, I=insertion on another image.
    }
    \label{tab:tab_penalties_combined}
\end{table}

%
%
%
%
%

\begin{figure}
    \centering
    \begin{tabular}{cc@{}c}
 \multicolumn{3}{c}{overlay\_onto\_screenshot intensity 93.8} \\
    \discimage{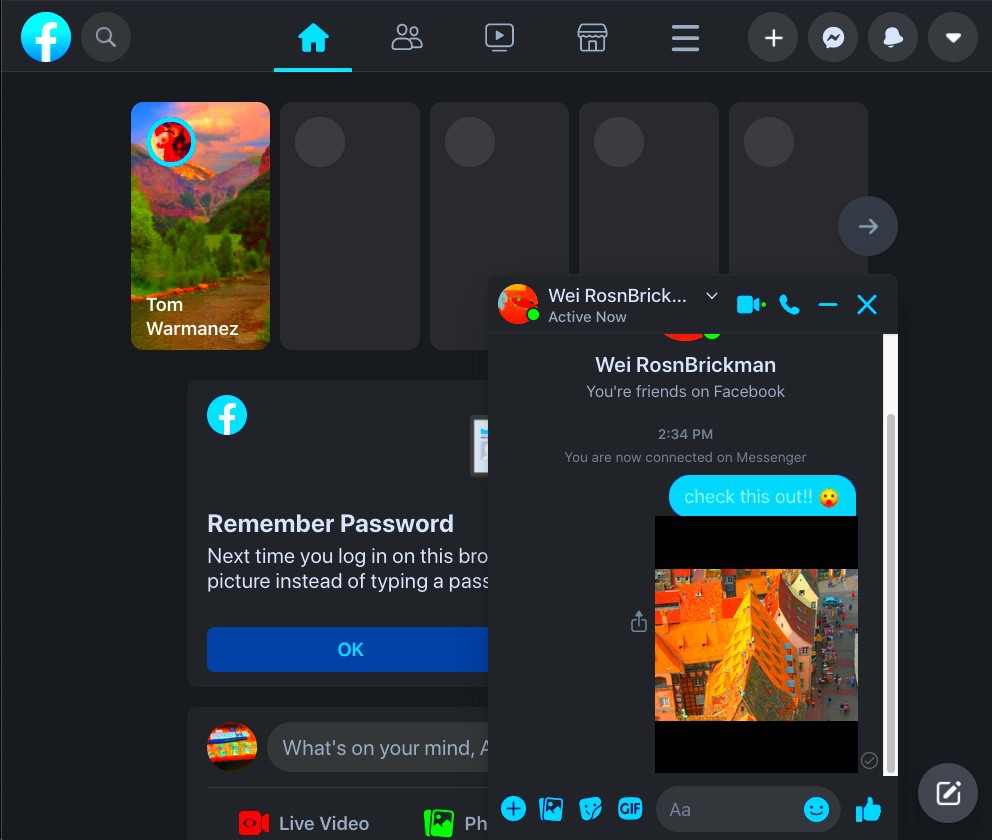} & \discimage{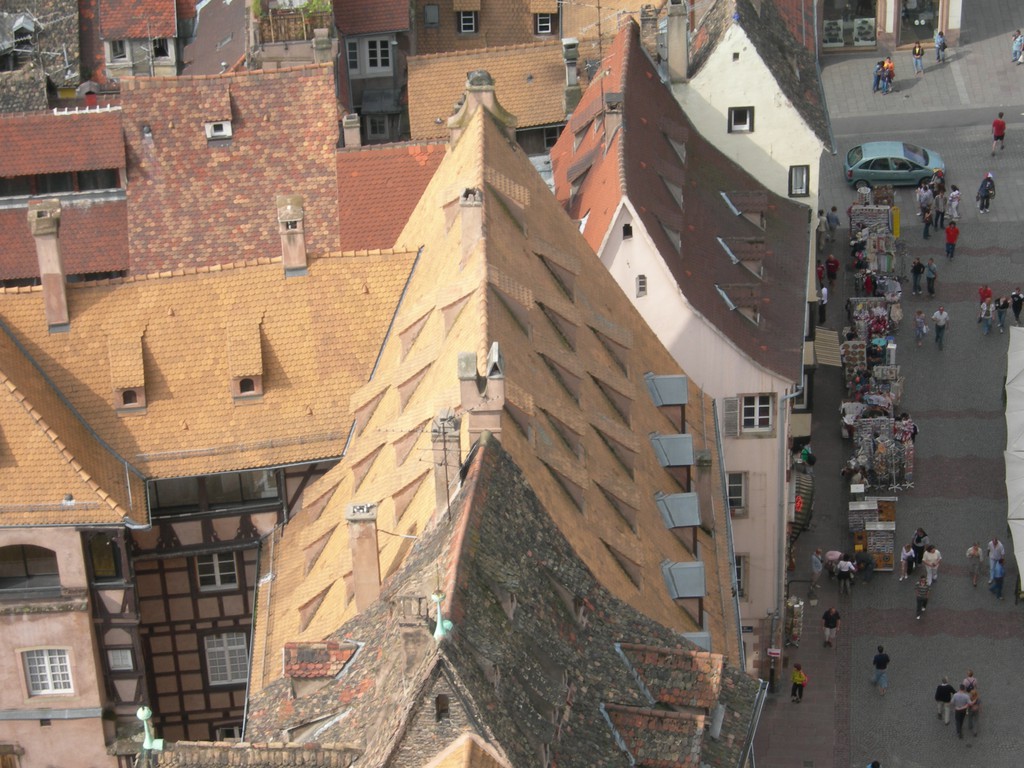} & \imcredsF{febs} \\    
\multicolumn{3}{c}{overlay\_burred\_mask intensity 90.6} \\
    \discimage{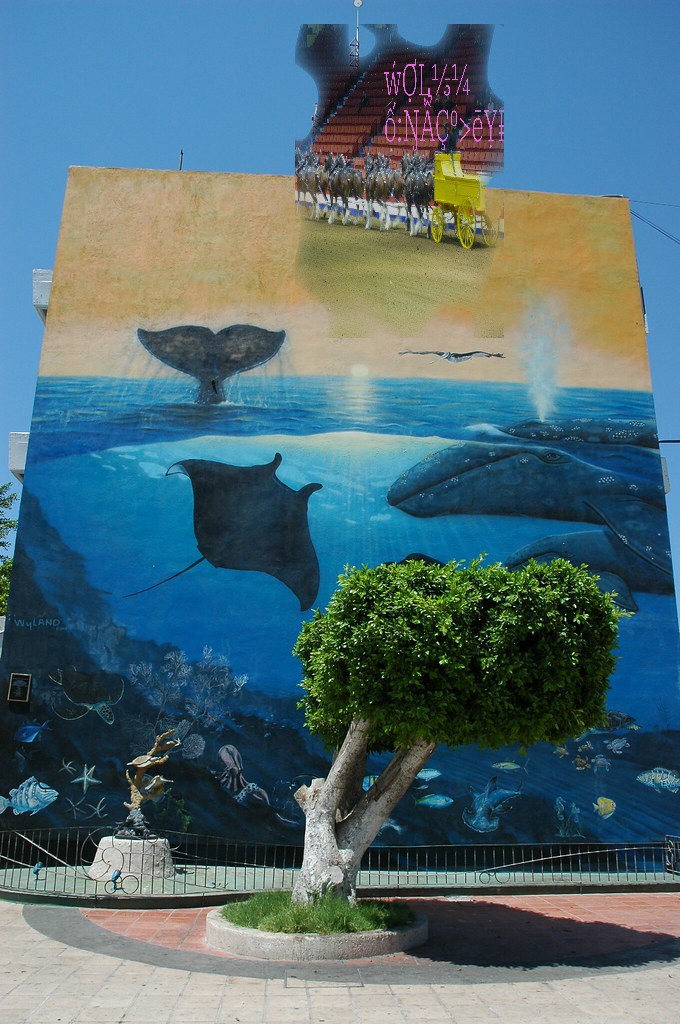} & \discimage{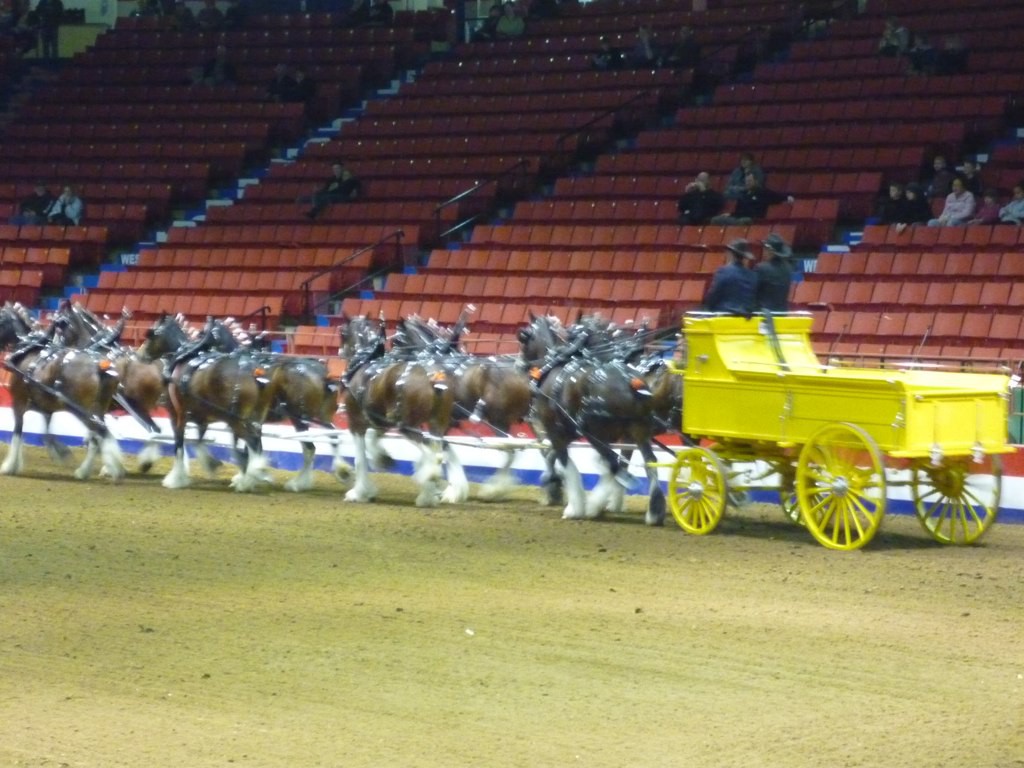} & \imcredsF{qwekiop147} \\
\end{tabular}%
\begin{tabular}{|cc@{}c}
 \multicolumn{3}{|c}{adversarial attack, PSNR=20, model resnet18} \\
    \discimage{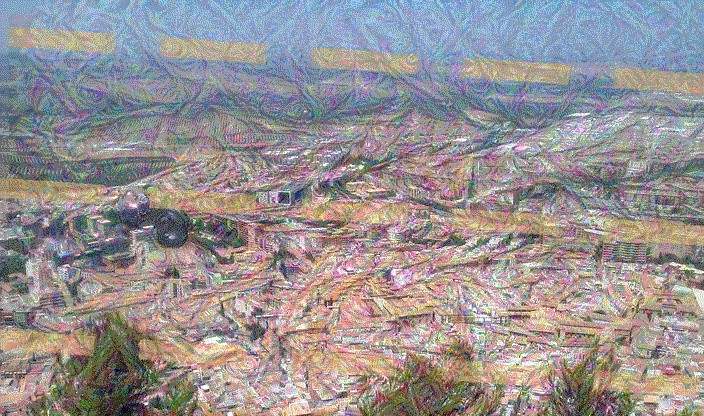} & \discimage{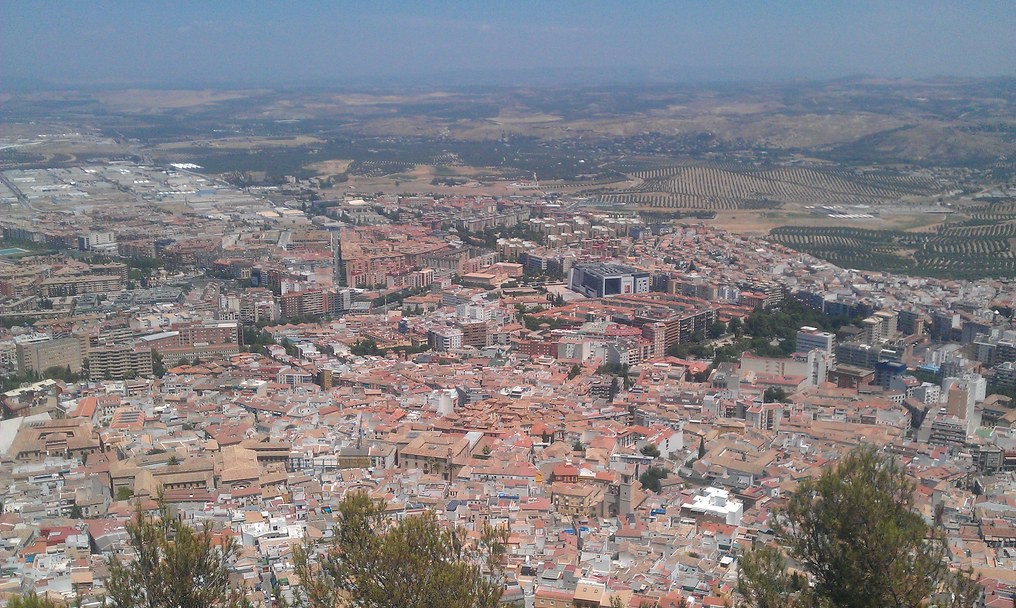} & \imcredsF{Francisco} \\    
 \multicolumn{3}{|c}{adversarial attack, PSNR=20, model dino} \\
    \discimage{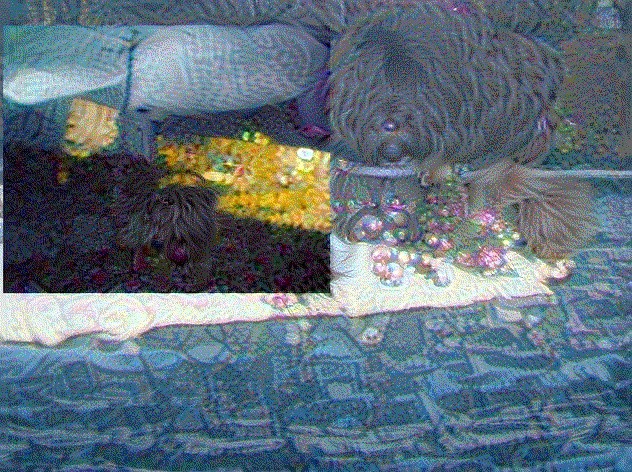} & \discimage{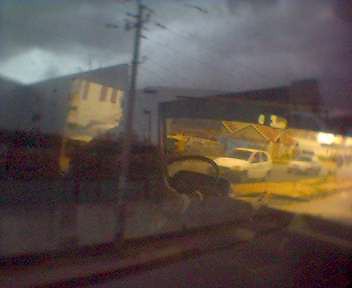} & \imcredsF{bbb\_js2000} \\
    \end{tabular}
    \caption{
    Example images with overlays and adversarial attacks.
    For each example the query is on the left, the reference image on the right.
    }
    \label{fig:example_over_aa}
\end{figure}

\subsection{The crop and overlay transformations}

The crop and overlay transformations are the hardest ones to handle, even in the matching track. 
Here we analyze how much of the image surface can be removed. 

Figure~\ref{fig:crop_overlay} shows the mAP for those transformations. 
The plot on the left shows that \teamname{VisionForce} submission is able to retrieve an image half of the cases even if only 6-12\% of the original image surface remains. 
On the right it appears that if the source image is inserted onto another and covers less than 20\% of the image surface, the  \teamname{VisionForce} method can recover it 80\% of the time.

\begin{figure}
    \includegraphics[width=0.45\linewidth]{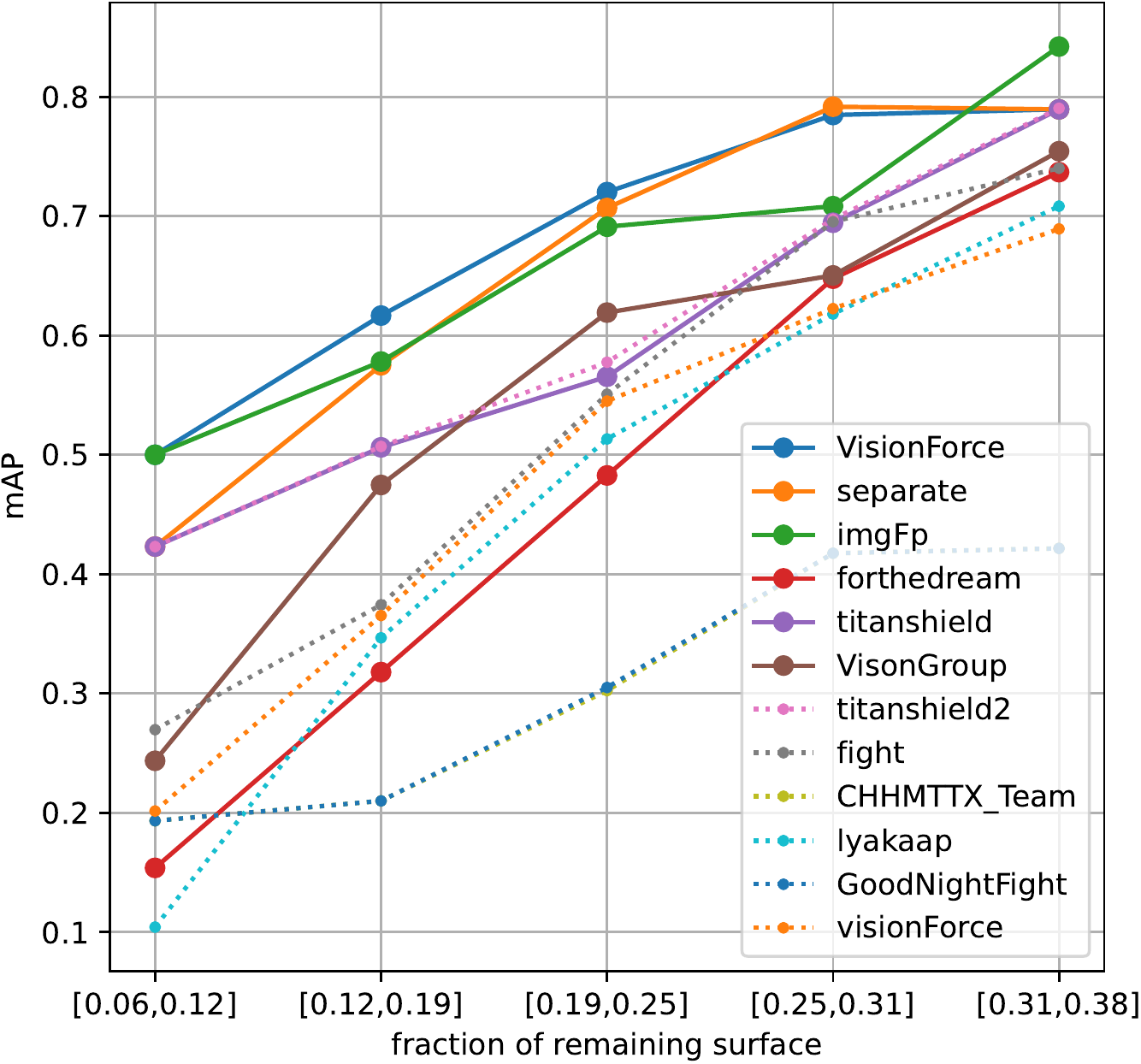}%
    \includegraphics[width=0.45\linewidth]{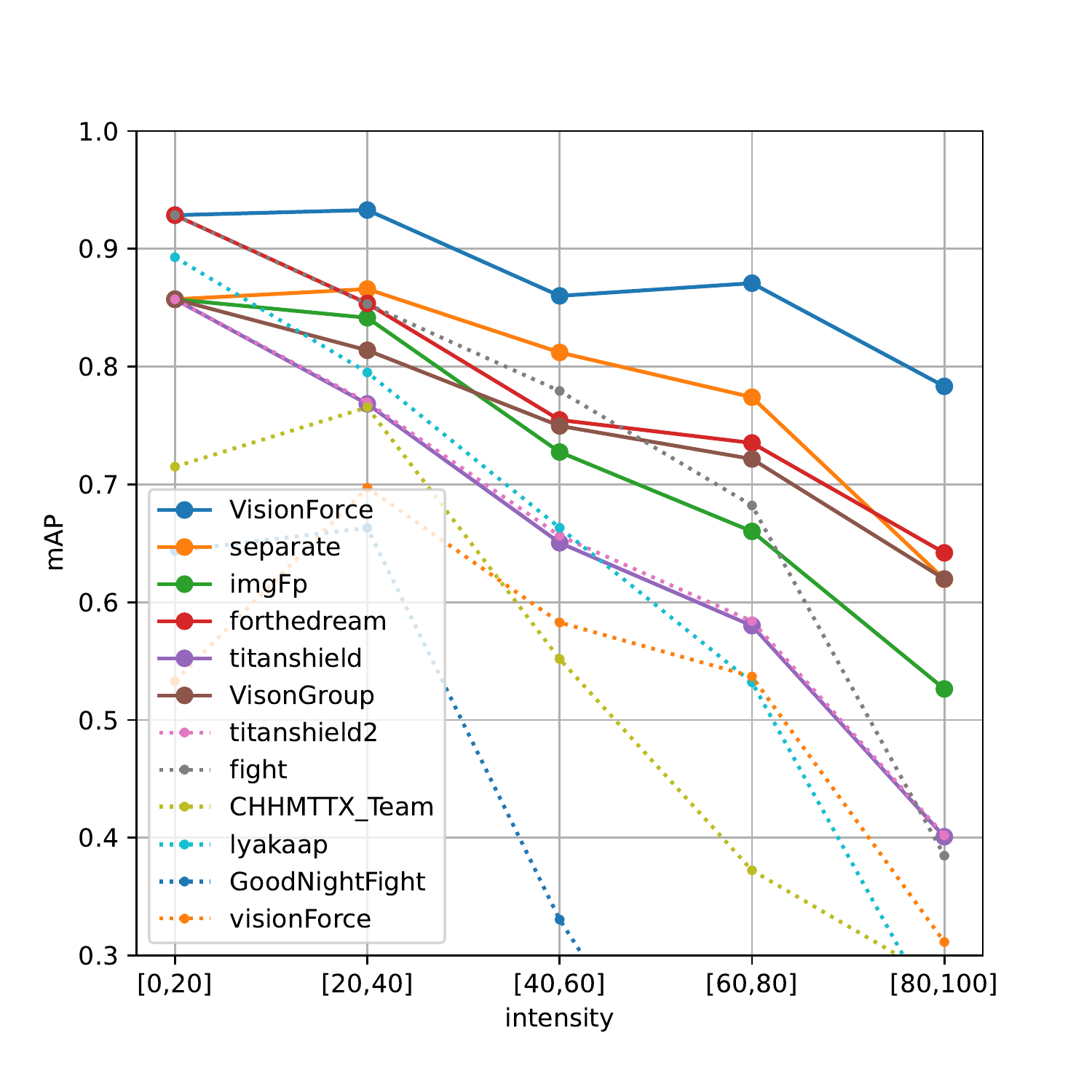}
    \caption{
        Retrieval performance (mAP) as a function of the intensity of some transformations that change the visible area of the source image.
        Descriptor track submissions are in dotted lines.
        Left: fraction of remaining surface for the cropping transformation. 
        Right: intensity of the overlay transformations, \ie the fraction of image pixels that does \emph{not} come from the source image. 
    }
    \label{fig:crop_overlay}
\end{figure}

\subsection{The adversarial augmentation transformation}

The adversarial augmentations were introduced in the final phase. 
The attack is effective only on the same model that it was trained on. 
Therefore it is interesting to look at the impact of the transformations as a function of their basis model.
Table~\ref{tab:tab_per_aa_model} shows this analysis. 
A few submissions are very sensitive to attacks based on the Dino model~\citep{caron2021emerging}, \eg submissions from \teamname{VisionGroup}, \teamname{mmcf} in the matching track. 
The SSCD~\citep{pizzi2022sscd} model is also quite effective. 
Both models were trained in a self-supervised way, and in addition SSCD was trained on the DISC training set. 

However, the overall analysis in Table~\ref{tab:tab_penalties_combined} shows that the adversarial augmentations are relatively harmless compared to other attacks. 
This is probably because they don't target the correct model exactly or because submissions use ensembles of models.

\begin{table}
\scalebox{0.6}{
\begin{tabular}{r|rrrrrrr}
 & \tcolhead{dino} & \tcolhead{resnet18} & \tcolhead{resnet34} & \tcolhead{resnet50} & \tcolhead{sscd} & \tcolhead{vgg16} & \tcolhead{vgg19}\\
 $n=$ & 172	& 157 & 161	& 172 & 176	& 176	& 187   \\
 \hline
VisionForce & \cellcolor[HTML]{fffa00}0.786 & \cellcolor[HTML]{c7ff00}0.830 & \cellcolor[HTML]{ffcf00}0.755 & \cellcolor[HTML]{d2ff00}0.823 & \cellcolor[HTML]{ffa400}0.727 & \cellcolor[HTML]{cdff00}0.827 & \cellcolor[HTML]{fffa00}0.786\\
separate & \cellcolor[HTML]{ff6e00}0.640 & \cellcolor[HTML]{9cff00}0.815 & \cellcolor[HTML]{ffda00}0.720 & \cellcolor[HTML]{d8ff00}0.773 & \cellcolor[HTML]{ff5d00}0.631 & \cellcolor[HTML]{f8ff00}0.750 & \cellcolor[HTML]{ffef00}0.733\\
imgFp & \cellcolor[HTML]{ffcf00}0.616 & \cellcolor[HTML]{8cff00}0.739 & \cellcolor[HTML]{ffe400}0.634 & \cellcolor[HTML]{cdff00}0.686 & \cellcolor[HTML]{ffe400}0.631 & \cellcolor[HTML]{d2ff00}0.682 & \cellcolor[HTML]{c7ff00}0.690\\
forthedream & \cellcolor[HTML]{ffa900}0.593 & \cellcolor[HTML]{d8ff00}0.683 & \cellcolor[HTML]{ffdf00}0.634 & \cellcolor[HTML]{fff500}0.650 & \cellcolor[HTML]{ffd400}0.625 & \cellcolor[HTML]{e8ff00}0.673 & \cellcolor[HTML]{f3ff00}0.663\\
titanshield & \cellcolor[HTML]{ffb400}0.587 & \cellcolor[HTML]{a2ff00}0.707 & \cellcolor[HTML]{fff500}0.634 & \cellcolor[HTML]{f8ff00}0.645 & \cellcolor[HTML]{ffc900}0.602 & \cellcolor[HTML]{ffef00}0.631 & \cellcolor[HTML]{d8ff00}0.668\\
VisonGroup & \cellcolor[HTML]{ff0028}0.422 & \cellcolor[HTML]{ffcf00}0.678 & \cellcolor[HTML]{ffa400}0.645 & \cellcolor[HTML]{ffdf00}0.692 & \cellcolor[HTML]{ff0008}0.523 & \cellcolor[HTML]{ffe400}0.692 & \cellcolor[HTML]{ffc900}0.674\\
mmcf & \cellcolor[HTML]{ff0028}0.446 & \cellcolor[HTML]{ffdf00}0.683 & \cellcolor[HTML]{ff8e00}0.625 & \cellcolor[HTML]{ffdf00}0.684 & \cellcolor[HTML]{ff1c00}0.545 & \cellcolor[HTML]{ffef00}0.696 & \cellcolor[HTML]{ffcf00}0.672\\
MeVer-QMUL & \cellcolor[HTML]{ff2700}0.410 & \cellcolor[HTML]{f3ff00}0.574 & \cellcolor[HTML]{edff00}0.576 & \cellcolor[HTML]{b2ff00}0.618 & \cellcolor[HTML]{ff7e00}0.471 & \cellcolor[HTML]{edff00}0.578 & \cellcolor[HTML]{ffdf00}0.543\\
VisonCognition & \cellcolor[HTML]{ff0023}0.429 & \cellcolor[HTML]{fff500}0.635 & \cellcolor[HTML]{ff8300}0.550 & \cellcolor[HTML]{fff500}0.631 & \cellcolor[HTML]{ff3d00}0.499 & \cellcolor[HTML]{fff500}0.632 & \cellcolor[HTML]{ffcf00}0.606\\
CHHMTTX & \cellcolor[HTML]{f3ff00}0.549 & \cellcolor[HTML]{8cff00}0.658 & \cellcolor[HTML]{d2ff00}0.571 & \cellcolor[HTML]{8cff00}0.622 & \cellcolor[HTML]{ddff00}0.563 & \cellcolor[HTML]{c7ff00}0.580 & \cellcolor[HTML]{e8ff00}0.556\\
\end{tabular}

\begin{tabular}{r|rrrrrrr}
 & \tcolhead{dino} & \tcolhead{resnet18} & \tcolhead{resnet34} & \tcolhead{resnet50} & \tcolhead{sscd} & \tcolhead{vgg16} & \tcolhead{vgg19}\\
 $n=$ & 172	& 157 & 161	& 172 & 176	& 176	& 187   \\
\hline 
titanshield2 & \cellcolor[HTML]{ffb400}0.588 & \cellcolor[HTML]{a2ff00}0.709 & \cellcolor[HTML]{fff500}0.634 & \cellcolor[HTML]{f8ff00}0.648 & \cellcolor[HTML]{ffc900}0.604 & \cellcolor[HTML]{fff500}0.636 & \cellcolor[HTML]{d8ff00}0.670\\
fight & \cellcolor[HTML]{ffb400}0.625 & \cellcolor[HTML]{9cff00}0.750 & \cellcolor[HTML]{e8ff00}0.692 & \cellcolor[HTML]{cdff00}0.712 & \cellcolor[HTML]{ff8900}0.594 & \cellcolor[HTML]{f8ff00}0.682 & \cellcolor[HTML]{cdff00}0.715\\
CHHMTTX\_Team & \cellcolor[HTML]{f3ff00}0.551 & \cellcolor[HTML]{8cff00}0.660 & \cellcolor[HTML]{d2ff00}0.573 & \cellcolor[HTML]{8cff00}0.622 & \cellcolor[HTML]{ddff00}0.564 & \cellcolor[HTML]{c7ff00}0.580 & \cellcolor[HTML]{e8ff00}0.556\\
lyakaap & \cellcolor[HTML]{ff7e00}0.481 & \cellcolor[HTML]{ffea00}0.560 & \cellcolor[HTML]{ffda00}0.548 & \cellcolor[HTML]{edff00}0.587 & \cellcolor[HTML]{ff3700}0.429 & \cellcolor[HTML]{ffda00}0.546 & \cellcolor[HTML]{ff8300}0.486\\
GoodNightFight & \cellcolor[HTML]{acff00}0.504 & \cellcolor[HTML]{8cff00}0.587 & \cellcolor[HTML]{cdff00}0.479 & \cellcolor[HTML]{8cff00}0.529 & \cellcolor[HTML]{cdff00}0.479 & \cellcolor[HTML]{b7ff00}0.495 & \cellcolor[HTML]{c2ff00}0.487\\
visionForce & \cellcolor[HTML]{ff0c00}0.364 & \cellcolor[HTML]{ddff00}0.563 & \cellcolor[HTML]{ff8900}0.455 & \cellcolor[HTML]{f8ff00}0.543 & \cellcolor[HTML]{ff3200}0.393 & \cellcolor[HTML]{ffc900}0.503 & \cellcolor[HTML]{ff9900}0.465\\
forthedream2 & \cellcolor[HTML]{ff8300}0.450 & \cellcolor[HTML]{b7ff00}0.590 & \cellcolor[HTML]{ffae00}0.480 & \cellcolor[HTML]{e3ff00}0.556 & \cellcolor[HTML]{ff8e00}0.456 & \cellcolor[HTML]{d2ff00}0.570 & \cellcolor[HTML]{ffa400}0.470\\
Zihao & \cellcolor[HTML]{ff2d00}0.327 & \cellcolor[HTML]{ffb400}0.425 & \cellcolor[HTML]{ff7800}0.380 & \cellcolor[HTML]{ffef00}0.467 & \cellcolor[HTML]{ff2700}0.322 & \cellcolor[HTML]{ff7e00}0.384 & \cellcolor[HTML]{ff6300}0.367\\
separate & \cellcolor[HTML]{ff4800}0.511 & \cellcolor[HTML]{ddff00}0.669 & \cellcolor[HTML]{e8ff00}0.662 & \cellcolor[HTML]{9cff00}0.715 & \cellcolor[HTML]{ff4800}0.513 & \cellcolor[HTML]{bdff00}0.693 & \cellcolor[HTML]{ffef00}0.632\\
S-square & \cellcolor[HTML]{ffa900}0.437 & \cellcolor[HTML]{d2ff00}0.528 & \cellcolor[HTML]{ffd400}0.468 & \cellcolor[HTML]{c7ff00}0.536 & \cellcolor[HTML]{ffae00}0.440 & \cellcolor[HTML]{f8ff00}0.500 & \cellcolor[HTML]{ffd400}0.468\\
\end{tabular}

}
    \caption{
    mAP for the adversarial attacks, broken down per attack model. Left: matching track, right: descriptor track.
    }
    \label{tab:tab_per_aa_model}
\end{table}

\ifarxiv
\newcommand{\mysubsection}[1]{\subsection{#1}}
\newcommand{\myparagraph}[1]{\paragraph{#1}}
\fi 

\ifnotarxiv
\newcommand{\mysubsection}[1]{\noindent\textbf{#1}}
\newcommand{\myparagraph}[1]{\emph{#1}}
\fi

\section{Top-ranked methods}
\label{sec:methods} 
The methods of top-ranked teams for the two tracks are presented. We refer to top-ranked methods simply as \emph{methods} and to top-participants as \emph{participants} in the following.
The major components that are common among different methods are identified and are used to structure this section; method details are provided per component while similarities and differences are discussed.
The top three methods, starting from the top-ranked for the matching and descriptor track are denoted by \mta~\citep{mt1}, \mtb~\citep{mt2},
\mtc~\citep{mt3}, and \dta~\citep{dt1}, \dtb~\citep{dt2}, and \dtc~\citep{dt3}, respectively.
\ifarxiv
Note that there were other competitive submissions, especially the \teamname{titanshield} submission that got the best performance on the descriptor track. %
However, the authors did not open source their method or publish a description. 
\fi

\mysubsection{Deep backbone and classical features}
All submissions rely on a neural net to analyze the images. 
We report the architecture here and discuss the training approach in the next subsection.

\mta uses all three  ResNet-50, ResNet-152, and ResNet50-IBN as backbones followed by GeM pooling~\citep{radenovic2018fine}, combined with WaveBlock~\citep{wang2022attentive}, and finally append a final projector module that consists of linear and non-liner layers and increases the dimensionality to 2048. WaveBlock can be seen as a type of augmentation method at the feature level. \dtc comes from the same team and uses the same backbones for both tracks.
\mtb uses ViT~\cite{dosovitskiy2021an} (``vit\_large\_patch16\_384'') to map an image to a global descriptor for the first ranking stage, and another ViT backbone (``vit\_large\_patch16\_224'') that receives an image pair in the form of a horizontally concatenated image as input and outputs a binary prediction for matching or non-matching image pair.
\mtc uses EsViT~\citep{li2021efficient} with Swin-B transformer~\citep{Liu_2021_ICCV}, adjusted as follows. 
Global average pooling is performed on the feature maps of each of the last blocks whose number of channels is $[512, 512, 1024, 1024]$, respectively. %
The output is concatenated and a fully connected layer is used to generate a 256-D global descriptor. %
This is the only method using classical local features too, in particular SIFT descriptors~\citep{lowe2004distinctive}.

\dta uses EfficientNetv2 with GeM pooling and reduce the dimensionality of the final descriptor by a linear layer with batch norm that is followed by l2 normalization. 
\dtb uses multiple backbones: EfficientNetV2 1, EfficientNetV2 s, EfficientNet b5, and NfNet 11. Each backbone is followed by GeM pooling~\citep{radenovic2018fine} and a linear layer to reduce the dimensionality and L2 normalization.

\mysubsection{Training approaches}

There were various training approaches for the given architectures, often decomposed into several phases that we describe here.

\myparagraph{Pre-training on external data} 
External datasets are used by all participants in the pre-training stage: either an existing pre-trained network is used or the participants performed the pre-training themselves. 
ImageNet is used in all cases, with supervised learning for \mtb, \dta, and \dtb, and with unsupervised learning for \mta, \mtc, and \dtc. 

\myparagraph{Training augmentations}
All methods compose an augmentation set that is richer than the conventional augmentation used to train classifiers, with transformations that mimic the task of copy detection. 
Such examples are more extreme geometric and photometric transformations and image/text/emoji overlay. 
\mta validates the impact of the enriched augmentations which is quantified to be significant. 
This is not surprising, as the copy detection task is close to self-supervised learning, where data augmentation is the only source of intra-class variablity~\citep{dosovitskiy2014discriminative,chen2020simple}.

\myparagraph{Training on ISC training set} 
\label{sec:isctrain}
The main training is performed on the provided training set by using augmentation in order to mimic the query attacks. This step is performed in a self-supervised way for all participants since the training set is not labeled. 
This process follows the concept of instance discrimination~\citep{wu2018unsupervised} where each image in the training set forms its own class, and any of its augmentations belongs to that class.
If not otherwise mentioned, the training optimizes a backbone network to generate a global image descriptor.

Deep metric learning is used by \mta with a combined classification and triplet loss, for which hard samples are mined.
\mtb trains with SimCLR~\citep{chen2020simple} where an augmented image is matched to the original one using InfoNCE loss. 
Similarly, \mtc uses a triplet loss. 
\dta uses a contrastive loss and cross-batch memory~\citep{wang2020xbm} where one augmentation of the training image is performed with the enriched augmentation set and the other augmentation of the same image with the conventional augmentation set.
\dtb uses ArcFace loss~\citep{deng2019arcface} and additionally combine ImageNet with the ISC training set in this step. The large output space raises challenges in the training, which is handled by  gradually increasing the number of classes in the training. 
\mtc uses triplet loss with hard-negative mining combined with cross-entropy loss. 

\myparagraph{Fine-tuning on ISC query/reference set}
The competition rules allow training using the provided labels in Phase I, \ie ground-truth that defines the correspondences between the provided queries that are not distractors and the reference images. Only \mtc and \dta perform such a fine-tuning process, which is shown to noticeably boost the performance. All other participants rely on their own augmentations applied to the training images in order to mimic query transformations, as described in Section~\ref{sec:isctrain}.

\mysubsection{Sub-image region detection and feature extraction}
Detecting regions appears to be important for the matching track.
\mta uses a fixed set of crops, regions detected by Selective Search~\citep{uijlings2013selective}, and regions detected by YOLOv5~\citep{yolov5} trained to detect overlays of other images or emojis. Overlays of the former are used in further processing, while overlays of the latter are ignored.
\mtc trains a pasted-image detector to obtain crops of possibly overlaid images during inference. Similarly to the approach of \mta, positive examples are synthetically created, but standard uninformative overlaid images such as emojis are considered negatives.

Only \dtc uses region detection for the descriptor track; \ie the same YOLOv5 is used as in \mta.

\myparagraph{Feature extraction}
\mta feeds the whole image but also each crop to the backbone and obtains a descriptor per case. Note that 33 backbones are used for the whole image, but only 3 of them are used for the region descriptors. This is done both for query and reference images. \mtb feeds the whole image or the concatenated one to ViT. \mtc use the whole reference and query image as input to the backbone, and additionally the region, if any, provided by the overlay detector on the query image. Moreover, SIFT is used for local feature detection and descriptor extraction on all images.

Methods for the descriptor track feed the whole input image to the backbone and obtain a global descriptor. An exception is \dtc who replace the full image with the region, if any, that is obtained with the Yolo-based overlay detector. 

\mysubsection{Ensembles}
Model and similarity combination is done in different ways. We point out the case of combining different backbones, \eg different architectures or multiple training runs of the same architecture, representation from fixed geometric augmentations performed at test time, or global and local representation.
Some methods use more than one of these ensemble types.

\myparagraph{Backbone ensemble} %
\dtb ensembles the representation of the different backbones by concatenation and dimensionality reduction with PCA. The backbones not only differ in terms of the architecture, but are also a result of training with a different number of training classes. In total, 7 backbones are ensembled.
\mta keeps the maximum similarity over 33 different backbones. 
The three previously mentioned backbone networks are pre-trained in a self-supervised way; each backbone is pre-trained with either ByOL~\citep{grill2020bootstrap} or Barlow-Twins~\citep{zbontar2021barlow}.
Additionally, each backbone is trained 11 times, each one with a different augmentation set. 
A special case, not directly fitting into this category, is \mtb who fuses the two ViT models, \ie the one for single image to obtain descriptor and the one for the concatenated image pair to obtain a relevance confidence. If the reference image is top-ranked with descriptor similarity, then relevance confidence is used to re-rank.

\myparagraph{Test-time augmentation ensemble} %
\dtc ensembles multi-resolution representations, simply by averaging and re-normalizing the descriptor obtained for input images at 4 different resolutions.
\mtc horizontally flips the query and maintains the maximum similarity over the two query versions.

\myparagraph{Global/local ensemble} %
Matching track  methods use an ensemble of global and local/regional processing.
\mta computes the similarity between the whole query image and each crop of a reference image and vice versa (reference versus query) and the maximum similarity is maintained. 
\mtc uses SIFT to estimate the \emph{SIFT-score} by counting the number of correspondences that are formed between query descriptors and the closest descriptor among all reference images whose similarity is above a certain threshold, and satisfy the ratio test~\citep{lowe2004distinctive}. The SIFT-score is used for all images that appear in the top similar images which are estimated in three different ways and then accumulated if an image appears in multiple top-ranked image shortlists. The three ways are (i) with CNN global descriptor from the full query, (ii) with CNN descriptor from the cropped query image according to the pasted-image detector, and (iii) with SIFT-score. The SIFT-score is estimated on the full query image in the first case, but in the detected region in the other two. In this way, information from the SIFT and CNN representations are fused.

\mysubsection{Score normalization} 
Score normalization is shown to be useful in the baselines provided with the DISC2021 dataset~\citep{douze20212021}. 
In particular, the similarity score is normalized \wrt the similarity between the query and images in the training set.
This approach is used by \mta in the matching track. All participants propose new ways to achieve a similar normalization in the descriptor track. Note that it is more challenging in that case, because any normalization needs to be applied a priori to the descriptor itself.
All three methods try to move the query or reference image descriptor far from descriptors of the training set. The performance impact of descriptor normalization is significant for all participants.

\mysubsection{Discussion}
\ifarxiv
A brief summary of the different method components is shown in Table~\ref{tab:methods_both}. %
\fi 
It turns out that top results are achieved with a variety of different approaches. The backbones that are used are either CNNs or ViTs; losses are either classification-based, pairwise-based, or both, while regional representation comes from fixed regions, trained detectors, or even SIFT. As common winning components we identify score normalization, strong augmentations that mimic image copies, and ensembles. Ensembles are a common winning component for research competitions without computational complexity constraints. Note that the top matching method relies on up to 33 different backbones and multiple image regions that are represented separately. The memory that is required to store the representation of all references images is around 900Gb, which is two orders of magnitude greater than the 1Gb needed for the global descriptor track approaches. Achieving high performance with limited resources is definitely a challenging task and interesting future direction.

\ifarxiv

\begin{table}[]
\small
    \centering
    \begin{tabular}{llllll}
    \textbf{Matching track}
     & Backbone & Loss & Region det. & Ensemble &
     \\\hline \hline
    \mta & ResNet & CE+triplet & \begin{tabular}{@{}l@{}}fixed, SS,\\YOLOv5\end{tabular} & \begin{tabular}{@{}l@{}}backbones,\\  multi-resolution,\\local-global\end{tabular} &
    \\[10pt] \hline
    \mtb & ViT & CO\&BCE & - & 
    \begin{tabular}{@{}l@{}}singe-image, \\ pairwise\end{tabular} &
    \\[5pt] \hline
    \mtc & EsViT & triplet & YOLOv5  & 
    \begin{tabular}{@{}l@{}}horizontal flip,\\local-global\end{tabular} &
    \\\hline
    \multicolumn{6}{c}{~} \\
    \multicolumn{6}{c}{~} \\
    \textbf{Descriptor track}
 & Backbone & Loss & Region det. & Ensemble & Score norm. \\\hline \hline
    \dta & EffNetv2 & CO & - & - & subtract \\\hline
    \dtb & \begin{tabular}{@{}l@{}}EffNet(v2),\\NfNet\end{tabular} & ArcFace & - & backbones & rescale \\\hline
    \dtc & ResNet & CE+triplet & YOLOv5 & multi-resolution & rescale \\ \hline
    \end{tabular}
    \caption{Summary of methods for the two tracks. 
    CE+triplet: combination of cross-entropy and triplet loss. 
    CO\&BCE: two trained models -- one with contrastive loss, another with binary cross-entropy loss. ``fixed'':  pre-defined set of regions. SS: Selective Search.
    For descriptor normalization, negatives (from training set) are either subtracted (followed by l2-norm) or used to rescale the query descriptor.}
    \label{tab:methods_both}
\end{table}

\fi

\section{Conclusion}

We organized the Image Similarity Challenge with the intention to introduce a benchmark for image copy detection and to push the state of the art in this field. The solutions from participants were of high quality, some of which introduce interesting new research directions. The main ingredients for the top submissions were careful tuning of data augmentation at training time, score normalization, explicit overlay detection and local-to-global comparison. We hope that this competition will spur more progress in the field of image copy detection, using the benchmark of the DISC21 dataset.

\subsection*{Acknowledgements} 
Thanks to Meta, which funded the competition and the prizes.
We thank Driven Data and in particular Greg Lipstein, Jay Qi and Mike Schlauch for organizing the competition.

\bibliographystyle{plainnat}

\bibliography{biblio}

\end{document}